\documentclass[letterpaper]{article} %
\usepackage[preprint]{aaai2027}  %
\usepackage[hyphens]{url}  %
\usepackage{graphicx} %
\usepackage{natbib}  %
\usepackage{caption} %
\usepackage{algorithm}
\usepackage{algorithmic}

\usepackage{newfloat}
\usepackage{listings}
\DeclareCaptionStyle{ruled}{labelfont=normalfont,labelsep=colon,strut=off} %
\floatstyle{ruled}
\newfloat{listing}{tb}{lst}{}
\floatname{listing}{Listing}

\usepackage{amsmath}
\usepackage{amssymb}
\usepackage{booktabs}
\usepackage{booktabs}
\usepackage{multirow}
\usepackage{svg}
\usepackage{multibib}
\newcites{app}{Appendix References}
\title{A Dataset and Model for Imputing Water Surface Elevation on a Large and Extremely Sparse Spatiotemporal Graph}

\author {
    Ruben Cartuyvels\textsuperscript{\rm 1}\thanks{ruben.cartuyvels@esa.int},
    Karim Douch\textsuperscript{\rm 2, \rm 7},
    Gabriele Bertoli\textsuperscript{\rm 3, \rm 4},
    Mounia El Baz\textsuperscript{\rm 1},
    Artemis Vrettou\textsuperscript{\rm 7},
    Sébastien Lefèvre\textsuperscript{\rm 1, \rm 5, \rm 6},
    Diego Fernandez Prieto\textsuperscript{\rm 2}\\
}
\affiliations {
    \textsuperscript{\rm 1}European Space Agency, $\Phi$-lab,
    \textsuperscript{\rm 2}European Space Agency, Science Hub,
    \textsuperscript{\rm 3}University of Florence,
    \textsuperscript{\rm 4}Imperial College London,
    \textsuperscript{\rm 5}Université Bretagne Sud, IRISA,
    \textsuperscript{\rm 6}UiT -- The Arctic University of Norway,
    \textsuperscript{\rm 7}Serco Italia SpA, Rome, Italy
}

\begin{document}

\maketitle

\begin{abstract}
Continuous monitoring of water surface elevation across river networks is critical for flood forecasting, water resource management, and understanding the global water cycle. Yet, the scarcity of in situ gauges across much of the globe constrains the development of reliable modeling frameworks. Satellite altimetry has the potential to alleviate this problem but its use is currently hindered by sparse temporal coverage. To this end, we introduce \texttt{AmazonWSE}, a dataset for training and evaluating large-scale spatiotemporal graph imputation methods that integrates processed satellite altimetry measurements from a range of sources, including the recent wide-swath SWOT sensor. The dataset covers approximately 19K river sections in the Amazon river basin over 10 years (2016–2026), with in situ gauges held out for evaluation. Besides contributing a novel real-world use case with the potential for societal impact, \texttt{AmazonWSE} introduces significant technical challenges: with fewer than 1\% of sections observed per day, the dataset is far sparser than existing imputation benchmarks, and its directed acyclic river topology is both structurally different from and larger than graphs in existing datasets. We show that prior spatiotemporal graph imputation methods are not adapted to this topology, scale and sparsity, and propose a simple bidirectional selective state space model that outperforms them by sampling connected subgraphs and flattening space and time into a single token sequence with topology-aware positional encodings. Compared to the state-of-the-art published method for SWOT-based WSE densification, which integrates statistics with physical modeling, our model reduces RMSE against in situ gauges by 18-39\%, while producing predictions for every river section rather than only those with sufficient nearby satellite coverage.
\end{abstract}

\section{Introduction}

\begin{figure}[ht!]
\centering
\includegraphics[width=0.999\linewidth]{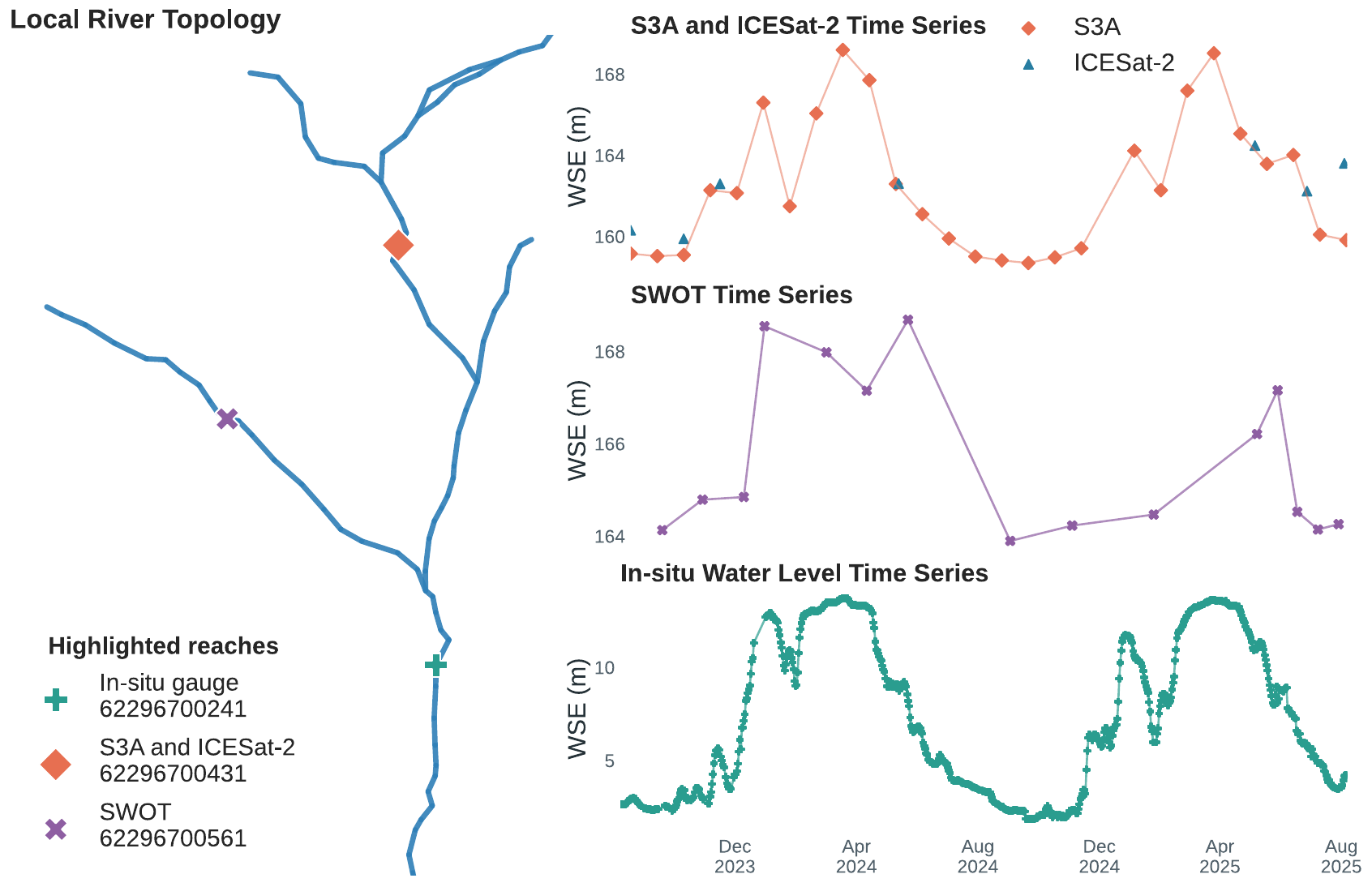}
\caption{Example time series and local graph topology in \texttt{AmazonWSE}. Locations can be observed by zero, one (such as the location observed by SWOT) or more (such as the location observed by ICESat-2 and S3A) sources.
The satellite sources measure with reference to the approximate sea level, while the in situ gauges measure deviations from a chosen location-specific reference level.}
\label{fig:source-timeseries}
\end{figure}

Monitoring water surface elevation (WSE) across river networks is essential for flood forecasting, water resource management, and understanding the global water cycle. The Amazon basin, the largest river system on Earth, is of particular importance because of its role in global climate regulation and biodiversity~\citep{fassoni2021amazon}.

Observations of WSE come from diverse sensors that each cover the spatiotemporal domain only partially. In situ flow gauges, such as those operated by Brazil's National Water and Sanitation agency (ANA), provide sub-daily measurements at fixed locations but cover unevenly a small fraction of the river network. Satellite altimetry missions have measured WSE globally since the 1990s, but these instruments observe a given location only every 10--91 days along narrow ground tracks~\citep{abdalati2010icesat2,normandin2018altimetry}. The Surface Water and Ocean Topography (SWOT) mission, launched in 2022, provides for the first time wide-swath observations of river WSE with a 21-day repeat cycle~\citep{biancamaria2016swot}, yielding unprecedented spatial coverage but a short temporal record.

We synthesize these sensor observations into a benchmark dataset for spatiotemporal graph imputation of daily WSE that observes the Amazon river network.
The spatiotemporal graph formed by the Amazon and its tributaries is extremely large with over 19K nodes (or \textit{reaches}; river sections of $\sim$8-12~km as defined in \citep{altenau2021sword}) and highly sparse: on any given day, fewer than 1\% of nodes carry an observation. 
Spatiotemporal graph neural networks (STGNN) for imputation
often take full graphs as input, which for large graphs is resource intensive.
Consequently, current methods are evaluated on small graphs of $\sim$1{,}000 nodes or less.
Furthermore, under extreme sparsity, same-day spatial neighborhoods contain almost no observed nodes, either starving the message-passing mechanism or introducing prohibitive amounts of virtual tokens.
Finally, we incorporate multiple sources that may observe the same node on the same day, yet due to different sensor characteristics, yield a different measurement.
Figure~\ref{fig:source-timeseries} shows example time series and a fragment of the underlying topology.

We show that recent transductive and inductive graph or recurrence based imputation methods do not perform well in our setting.
We introduce a simple sequence model
which flattens space and time into a single token sequence of observed measurements.
This model handles the sparsity more naturally than grid-based STGNNs and its inductive bias fits the sequential upstream$\to$downstream and temporal order inherent to river systems. 
Our results support recent findings that GNNs applied to river networks for discharge prediction show limited benefit from graph topology~\citep{kirschstein2024river}. 
Importantly, we find that river topology \emph{does} inform better predictions, but is better exploited through subgraph sampling and metadata encodings than through explicit graph convolutions.

In summary, this paper makes the following contributions:
\begin{enumerate}
    \item We introduce a dataset of Amazon WSE time series on an underlying river 
    graph 
    that observes 19K river sections with a daily frequency over 2016--2026 and serves as a benchmark for imputation under extreme sparsity. 
    \item We propose a Mamba-based sequence model as baseline that we train for masked reconstruction on the heterogeneous observations with topology-aware subgraph samples~\cite{gu2023mamba}.
    \item We show that this model outperforms
    interpolation and existing transductive and inductive neural methods,
    and provide ablation studies.
    We demonstrate improvements in coverage and accuracy against Reach-Reg~\citep{halicki2026reachreg}, a current state-of-the-art method for densification of altimetry-derived WSE from SWOT data.
\end{enumerate}

\section{Related Work}

\begin{table}[t]
\centering
\small
\begin{tabular}{lrrrl}
\toprule
\textbf{Dataset} & \textbf{Nodes} & \textbf{Obs.} & \textbf{Missing} & \textbf{Graph} \\
\midrule
AQI-36             & 36    & 274K  & 13.24\% & Cities \\
METR-LA            & 207   & 6.52M   & 8.11\%  & Road \\
PEMS-BAY           & 325   & 16.94M  & 0.003\% & Road \\
CausalRivers  & 666   & 107M & ${\sim}$8\% & River \\
LamaH-CE & 859 & 11B & <20\% & River \\
LargeST-CA         & 8.6K  & 4.52B & n.r.    & Road \\
\midrule
\texttt{AmazonWSE} & 19.2K & 1.9M    & 99\%  & River \\
\bottomrule
\end{tabular}
\caption{Spatiotemporal dataset comparison. \textbf{Obs.}: measurements, reported or estimated from time steps and sparsity.}
\label{tab:dataset_comparison}
\end{table}

\begin{table}[t]
\centering
\small
\begin{tabular}{lrrrr}
\toprule
\textbf{Source} & \textbf{Nodes} & \textbf{\# obs.} & \textbf{Period} & \textbf{\% obs.} \\
\midrule
SWORD reaches   & 19,172 & -- (static) & -- \\
SWOT RiverSP    & 10K & 353K  & 2023--26 & 1.69\% \\
HydroWeb        & 3.8K  & 391K & 2016--26 & 0.50\% \\
ICESat-2        & 18K & 208K  & 2018--26 & 0.32\% \\
\midrule
Total observed & 18.5K & 1.9M & 2016--26 & 0.97\%\\
\textit{In situ (eval)}     & 375    & 955K & 2016--26 & 1.32\% \\
\bottomrule
\end{tabular}
\caption{Dataset statistics for the Amazon basin. Percentage observed (\textbf{\% obs.}) is calculated over the respective periods of operation of the sources.}
\label{tab:dataset_stats}
\end{table}

\paragraph{Satellite Altimetry for River Monitoring.}
Classical altimetry satellites provide an estimate of the WSE along their ground-track by measuring the distance between the satellite and the water surface with a radar,
and have provided observations since 1991. 
The HydroWeb database aggregates measurements from ERS-1/2, Topex/Poseidon, Jason-1/2/3, Envisat, Saral/AltiKa, Sentinel-3A/B, and Sentinel-6A into 
time series at locations where the satellite ground track crosses a river~\cite{santosdasilva2010water,normandin2018altimetry}. 
While other altimeters observe only along their precise ground track, the novel wide-swath InSAR sensor of the SWOT satellite observes entire contiguous river segments rather than point crossings~\citep{biancamaria2016swot}. 
Reach-Reg~\citep{halicki2026reachreg} exploits the SWOT measurement geometry for spatiotemporal WSE densification by chaining linear regressions between simultaneously observed 
reaches and by modeling water velocity.
They achieve the best accuracy with
re-processed SWOT data which is not available at scale~\cite{schwatke2015dahiti} but show that their method also works on public SWOT RiverSP data. 
However, like other works~\cite{tourian2016spatiotemporal,nielsen2022river}, they consider only large and well observed rivers without complex topology.

\paragraph{Spatiotemporal Imputation.}
Early work focused on modeling the temporal dimension~\citep{yi2016stmvl,cao2018brits}. 
Subsequent works model spatial interactions more explicitly and can be categorized into transductive methods, that assume to see every to-be-predicted node during training~\cite{cini2022grin,marisca2022spin,liu2023pristi,cheng2024faststi,de2024ggnet,nie2024imputeformer,yang2025gsli} and inductive methods, that transfer to entirely unseen nodes~\cite{wu2021ignnk,zheng2023increase,li2025stagann,xu2025kits,ren2026anchorgk,liang2026darkfarseer}.
Even though some methods introduce mitigations for the cost of processing large graphs, none of these has been evaluated on graphs with more than 1{,}200 nodes. 
SPIN~\cite{marisca2022spin} and IGNNK~\cite{wu2021ignnk} support subgraph sampling, similar to the model proposed here.
Only SPIN~\cite{marisca2022spin} and ImputeFormer~\cite{nie2024imputeformer} have been shown to work (transductively) at sparsity levels exceeding 90\%, but not up to 99\%~\cite{marisca2022spin,nie2024imputeformer}.
Forecasting methods exist for larger but denser graphs~\cite{cini2023sgp,liu2023largest}.

\begin{figure*}[t]
\centering
\includegraphics[width=0.198\linewidth]{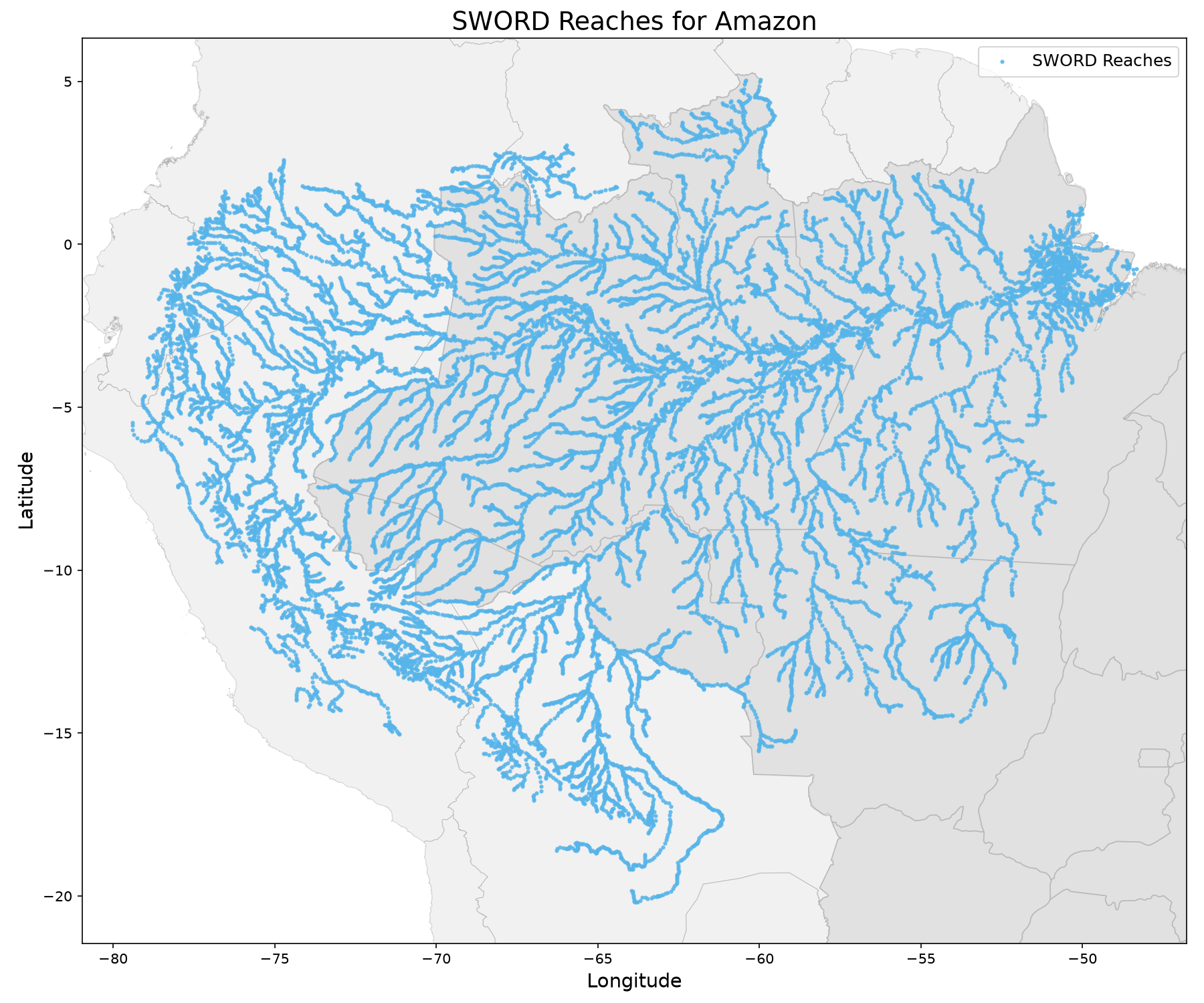}\hfill
\includegraphics[width=0.198\linewidth]{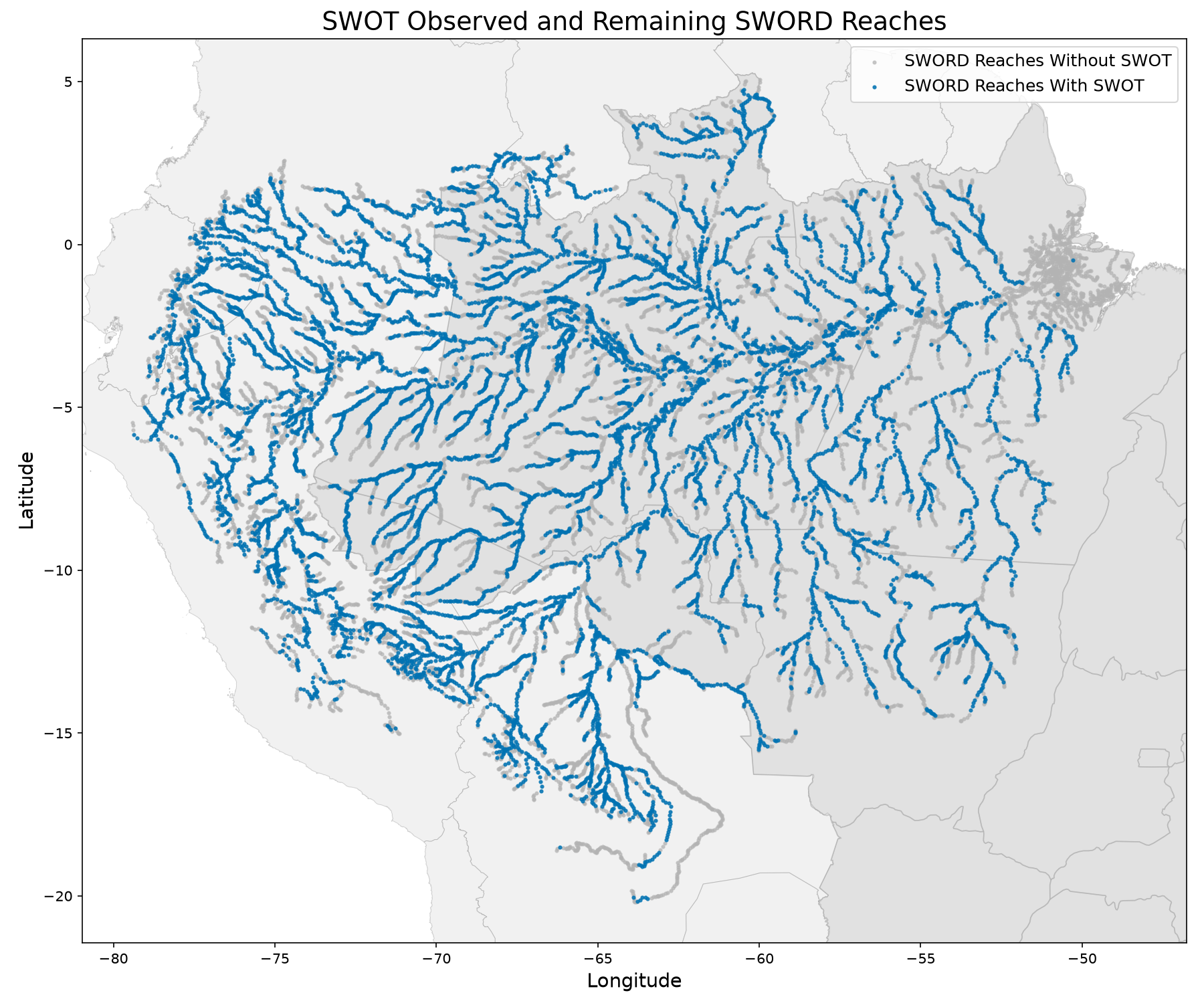}\hfill
\includegraphics[width=0.198\linewidth]{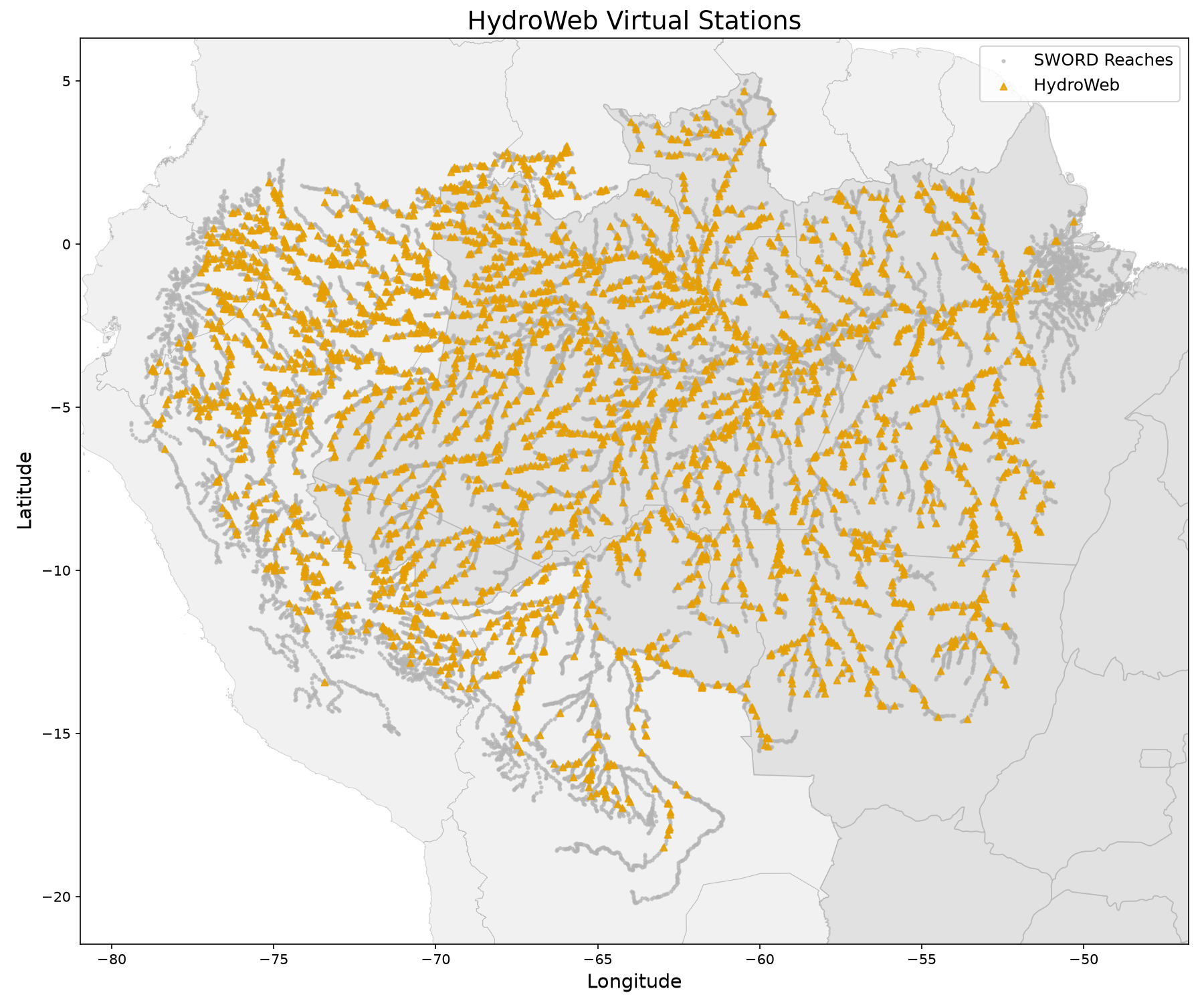}\hfill
\includegraphics[width=0.198\linewidth]{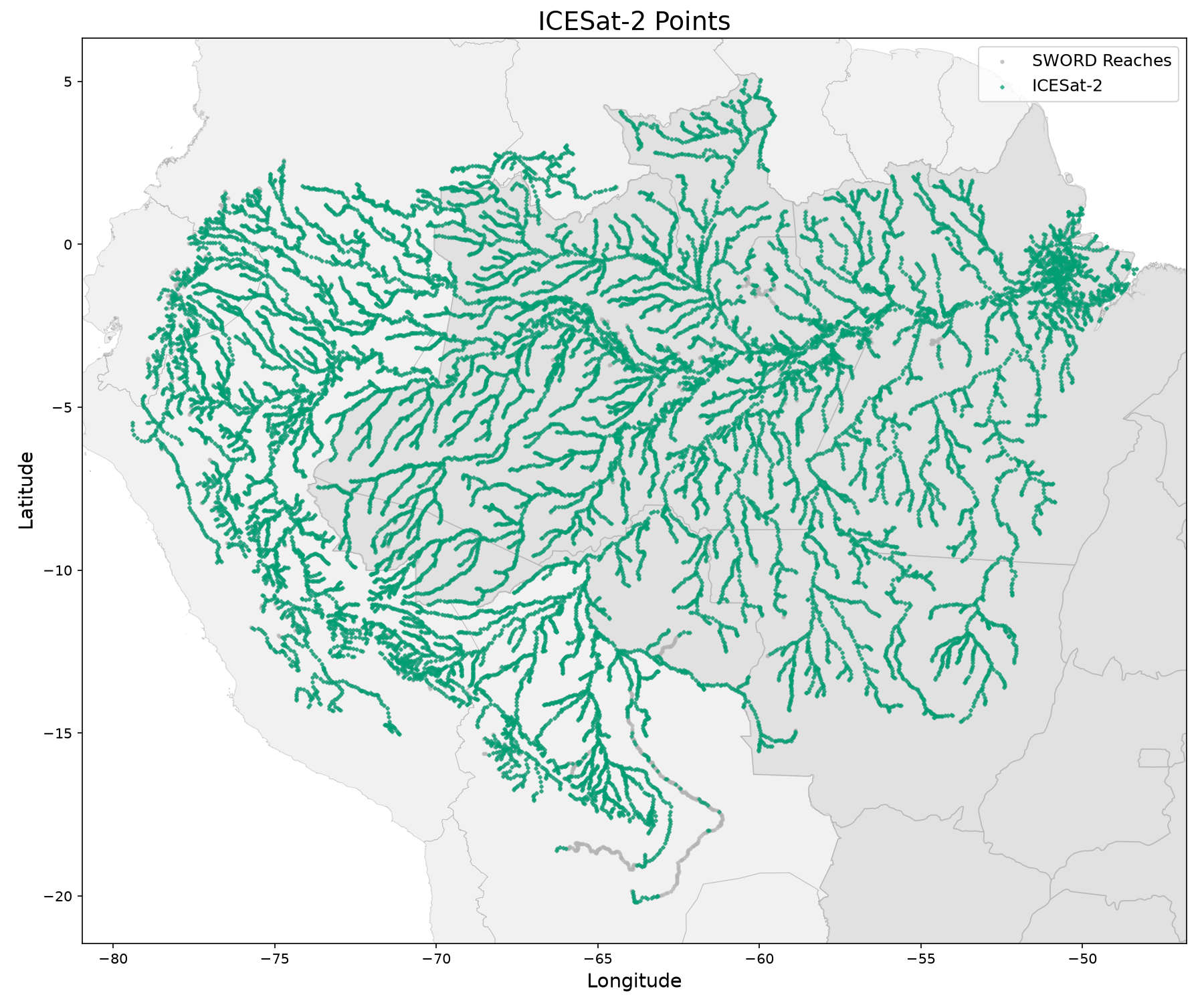}\hfill
\includegraphics[width=0.198\linewidth]{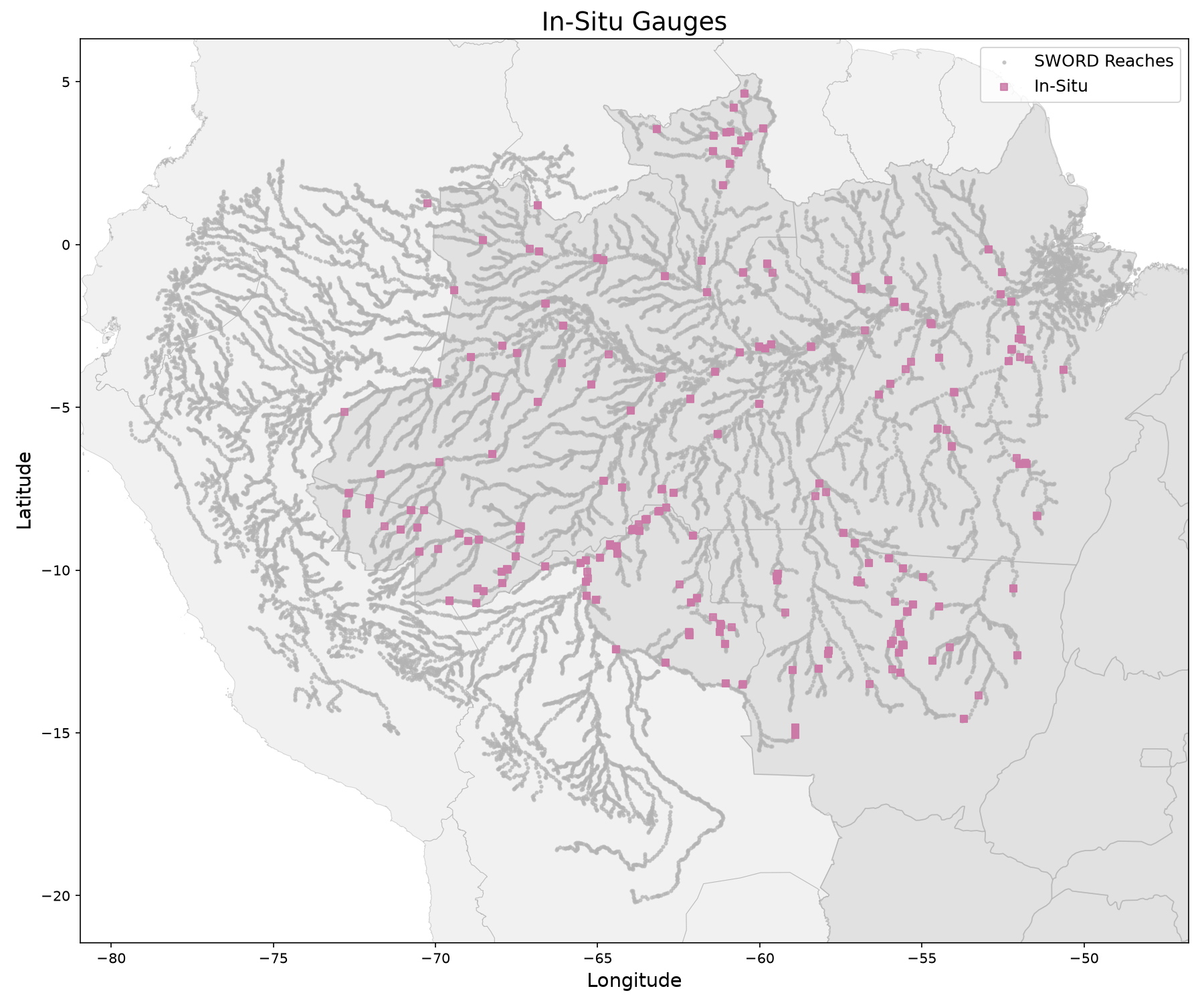}
\caption{Spatial distribution of data sources in \texttt{AmazonWSE}, after quality filters, with SWORD river topology background. Left to right: SWORD reaches, SWOT-observed reaches, HydroWeb virtual stations, ICESat-2 transects, ANA in situ gauges.}
\label{fig:data_sources}
\end{figure*}

\paragraph{Existing Datasets.} 
Benchmarks used for spatiotemporal imputation, though having more observations along their temporal axis, are substantially smaller in their spatial extent than \texttt{AmazonWSE} (Table~\ref{tab:dataset_comparison}). 
AQI-36 contains 36 hourly air-quality series, METR-LA and PEMS-BAY contain 207 and 325 five-minute traffic series, respectively~\cite{zheng2013aqi36,li2018dcrnn}. 
LargeST scales traffic forecasting to 8,600 sensors over five years, and CER-E and PV-US contain energy grid time series over $\sim$5{,}000-6{,}000 nodes, but none of these has been used for imputation~\citep{commission2012cer,hummon2012sub,liu2023largest}.
Table~\ref{tab:dataset_comparison} shows that these datasets offer denser data, which hampers the comparability of imputation studies that report incompatible artificial sparsification scenarios, from random missing entries~\cite{nie2024imputeformer} and temporal/spatial blocks~\cite{liu2023pristi} to entirely withheld target nodes~\cite{yang2025drik}, removed context nodes~\cite{zheng2023increase}, and held-out variable channels~\cite{de2024ggnet}.
\texttt{AmazonWSE} offers a consistent benchmark for extreme sparsity; by combining 1.9M irregular observations at 19K nodes over 2016--2026, fewer than 1\% of reaches are observed on any day. 
Its directed acyclic topology is also novel with respect to the cyclical graphs of existing datasets.

Hydrology datasets with discharge or water level time series aggregate data spatially into catchments and thereby erase the topology~\cite[Caravan;][]{kratzert2023caravan}, or only make available in situ gauge data, no satellite altimetry: LamaH-CE;~\cite{klingler2021lamah} and CausalRivers~\citep{causalrivers2025}.

\paragraph{Rivers and GNNs.}
\citet{acosta2025dualfloodgnn,taghizadeh2025hydrographnet} propose GNNs for spatial flood forecasting. 
\citet{dufourg2024forecasting} find that a GNN slightly outperforms (Conv)LSTMs when forecasting a water index from satellite image time series. 
\citet{kirschstein2024river}, on the other hand, report that reflecting river network topology through node adjacency in GNNs does not improve discharge forecasting over using isolated node-level MLPs. 
Here, we find that topology information does improve WSE reconstruction if used to sample context and encoded through positional encodings rather than through GNN message passing.

\section{Dataset}

We construct a multi-source dataset of water surface elevation time series for the Amazon river by integrating five data sources, each with different spatial and temporal characteristics. Figure~\ref{fig:data_sources} shows the spatial distribution of the data sources. Table~\ref{tab:dataset_stats} summarizes the resulting dataset.

\subsection{SWORD and Topology}

The SWORD database (v17b) was designed as a complementary static dataset to SWOT products and provides a topological graph of 19,172 river reaches in the Amazon basin \cite{altenau2021sword}.
Each reach is 
characterized by a unique ID, geographic coordinates, and upstream/downstream connectivity.
SWORD provides the graph structure, shown in Figures~\ref{fig:source-timeseries}-\ref{fig:subgraph}, on which our model operates and the static metadata used for spatial encodings.

The river topology defines a directed graph \(\mathcal{G}=(\mathcal{V},\mathcal{E})\), where \(\mathcal{V}\) is the set of SWORD reaches and \((u,v)\in \mathcal{E}\) indicates that \(v\) is immediately downstream of \(u\). 
The graph is acyclic, \(v\not\leadsto v\), and approximately but not strictly a tree, since reaches sometimes have multiple parents.

\subsection{Observation Sources}

\paragraph{SWOT RiverSP.}
The SWOT RiverSP product provides WSE observations mapped to SWORD reaches, from the wide-swath InSAR sensor \cite{swot}.
We use data from the science cycle that started in July, 2023. 
Of the 19,172 SWORD reaches, approximately 17K are observed by SWOT, and 10K are retained by quality filtering.
The repeat cycle takes 21 days, 
but due to the wide-swath sensor some reaches are observed several times during this period.

\paragraph{HydroWeb.next.}
The HydroWeb database provides WSE time series at 7.3K river crossings in the Amazon basin, derived from altimetry missions (including S3A, shown in Figure~\ref{fig:source-timeseries}) spanning 1993--2026~\citep{cretaux2015hydroweb}.
We use both the operational and research collections and retain 3.7K locations with data between 2016--2026 matched to a nearby reach.
Each virtual station has a measurement every 10--35 days depending on the altimetry mission.

\paragraph{ANA In Situ Gauges.}
The Brazilian National Water Agency (ANA) operates flow gauges across the Amazon basin. 
We retrieve WSE records via their API~\citep{ana_hidroweb_api}.
We retain 375 gauges with data between 2016--2026 after quality filtering, manual inspection and matching to SWORD reaches. 
Following standard practice in hydrology, we provide the gauge data as ground truth.
They provide dense temporal sampling (with an aggregated daily observation during periods of operation)
but are geographically sparse.

\paragraph{ICESat-2.}
The NASA Ice, Cloud and land Elevation Satellite-2 (ICESat-2) mission carries a laser altimetry instrument,
which measures along three pairs of narrow laser beams~\cite{abdalati2010icesat2}. 
We use the mean along-track surface elevation for each beam for each transect across a water body from the Level3B ATL22 Mean Inland Surface Water Data product~\cite{jasinski2025atl22}. 
Observations span Oct. 2018--Mar. 2026; with a nominal repeat cycle of 91 days and a large spatial coverage.
After filtering and matching, time series remain for $>$90\% of SWORD reaches.

\subsection{Processing and Quality Filtering}
\label{sec:quality-filtering}

HydroWeb, ANA, and ICESat-2 locations are assigned to the \textbf{nearest SWORD reach}, retaining only matches within 10~km (unmatched locations are discarded, imposing a stricter limit than the 20~km limit used by~\citet{halicki2026reachreg}).
SWOT RiverSP observations are already mapped to SWORD reaches. 
HydroWeb, SWOT, and ICESat-2 report elevations referenced to the
\textbf{EGM2008 geoid}~\cite{pavlis2012egm2008}, which provides a gravity-adjusted estimate of ``elevation above sea level''.
ANA records are in a local gauge
datum; for evaluation, each gauge is therefore mapped to
the prediction datum by a fitted linear transformation, following
\citet{halicki2026reachreg}.
Every source is placed on the common daily grid from 2016-01-01 to 2026-05-01 (3,774 days). 

Further processing is source-specific and according to expert hydrology standards. For \textbf{SWOT}, we 
screen product quality fields including reach quality, width,
cross-track distance, crossover calibration, random WSE
uncertainty, and severe bit flags. These criteria combine filters adapted from~\citet{andreadis2025swot} and~\citet{halicki2026reachreg} with additional harmonic-residual, observed-pixel precision and
$\Delta\text{width}/\Delta\text{WSE}$ consistency checks we introduce. 
\textbf{HydroWeb} text products are parsed from both operational
and research collections; invalid fill values and the short Jason-2 interleaved (J2N) record
are removed before daily alignment. \textbf{ICESat-2} transects over reservoirs and
transects shorter than 50~m are discarded, followed by a per-reach harmonic-residual
filter. \textbf{ANA} sub-daily measurements are reduced to a
daily median, and filtered using seasonal-trend residuals, followed by manual inspection by an expert.
More details are given in Appendix~\ref{app:preprocessing}.

All data (filtered and original measurements, uncertainties, quality flags, metadata) is provided in \texttt{h5netcdf} format and can easily be read by \texttt{xarray}.
Each dataset may be used freely for any purpose under its provider's terms;
some providers require attribution.
Appendix~\ref{app:storage-format} explains the storage format.
All data export, processing, loading and model code 
will be made public upon acceptance.\footnote{\url{https://github.com/rubencart/AmazonWSE}}

\subsection{Task: Spatiotemporal Imputation}
\label{sec:task}

Given whichever altimetry measurements are available, the task is to reconstruct
a daily WSE at each SWORD reach. 
The evaluation is broken down in two tracks: reconstruction
during the SWOT era (2023-07 to 2026-05), and a pre-SWOT hindcast. 
ANA gauges are not used as model inputs or as training targets but are reserved only for evaluation.

\begin{figure}[t]
\centering
\includegraphics[width=0.8\linewidth]{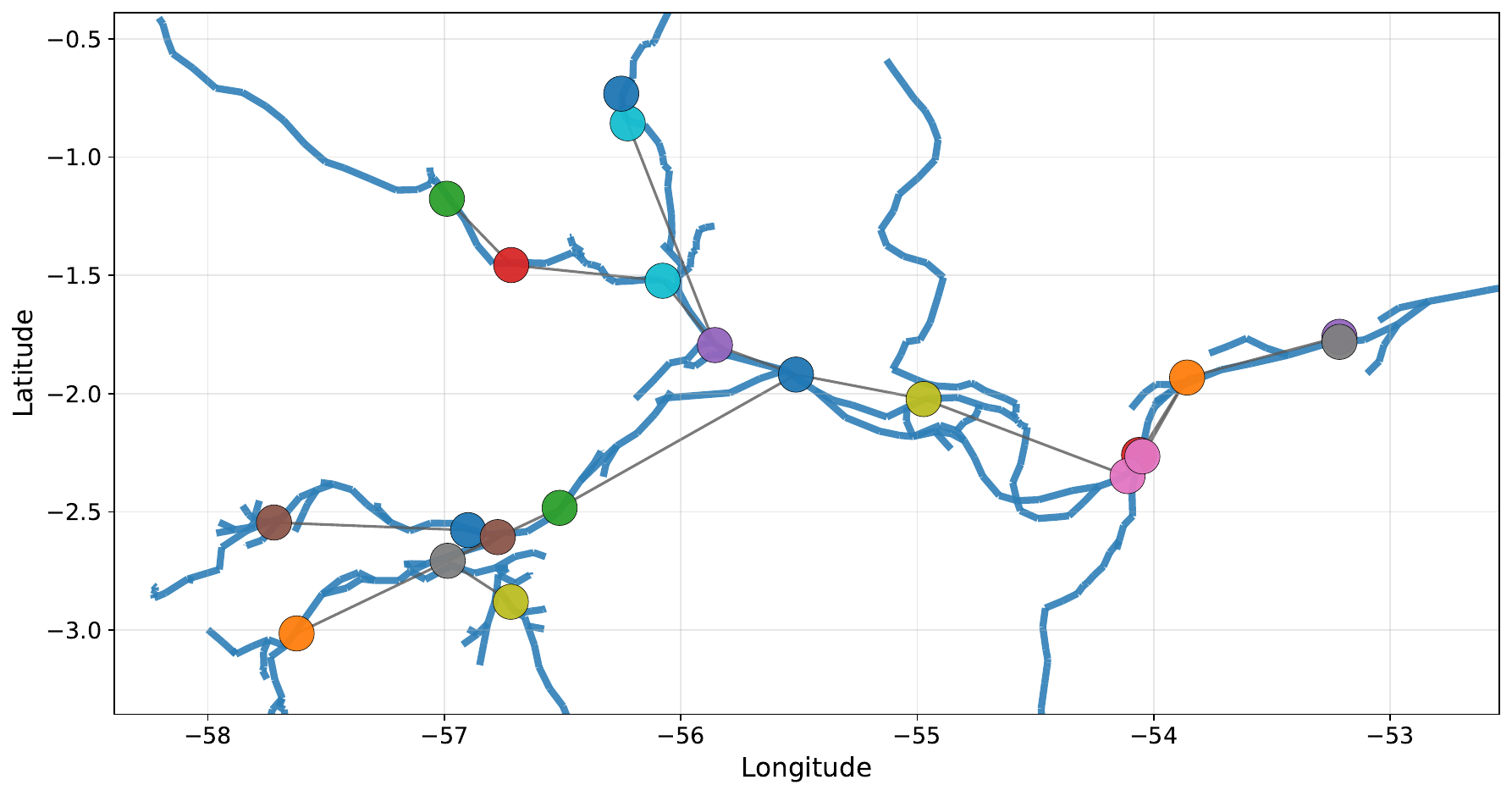}
\caption{Example of sampled subgraph (local edges in grey) with river topology (SWORD reaches in blue).}
\label{fig:subgraph}
\end{figure}

\section{Model}

We formulate WSE densification as masked reconstruction on the directed SWORD graph
$\mathcal{G}=(\mathcal{V},\mathcal{E})$. An observation is a tuple
$(j,i,t,r,h_{j,t}^{(r)})$, where source location $j$ is mapped to SWORD reach
$i\in\mathcal{V}$ by $i = \nu(j)$, $t$ is a daily date bin, $r$ identifies the observing source, and
$h_{j,t}^{(r)}$ is the measured WSE. Multiple sources may therefore produce distinct
tokens for the same reach and day.

\subsection{Sample Construction}

A sample is defined by an anchor location $a$, a contiguous time interval $I$, and a
connected local neighborhood $\mathcal{V}_a\subset\mathcal{V}$ obtained by following the
SWORD topology upstream and downstream from the anchor (Example in Figure~\ref{fig:subgraph}). Let $\mathcal{J}_a$ be the
selected source locations mapped to these reaches. The observed part of the sample is
\begin{align}
\mathcal{D}_{a,I}
=
\bigl\{(j,i,t,r):&
\, j\in\mathcal{J}_a,\ i=\nu(j),\ t\in I,\label{eq:sample-construction}\\
&\qquad h_{j,t}^{(r)}\ \text{is observed}\bigr\}.
\nonumber
\end{align}
We create tokens only for available measurements rather than constructing a
dense reach--time grid. Thus, sparsity reduces sequence length instead of filling the
input with missing-value tokens. 

\paragraph{Token Representation.}
For token $\ell$, let $z_\ell$ be the WSE normalized by the mean/std of the entire time series at source location $j$ (different sources observing the same location are normalized separately).
The embedded token $\mathbf{e}_\ell$ that is input to the model is a linear projection of $z_\ell$ if the token is not masked and a learned mask token embedding $\mathbf{e}_{\mathrm{mask}}$ if it is masked (for training) or a query token (for inference).

\begin{figure}[t]
\centering
\includegraphics[width=0.95\linewidth]{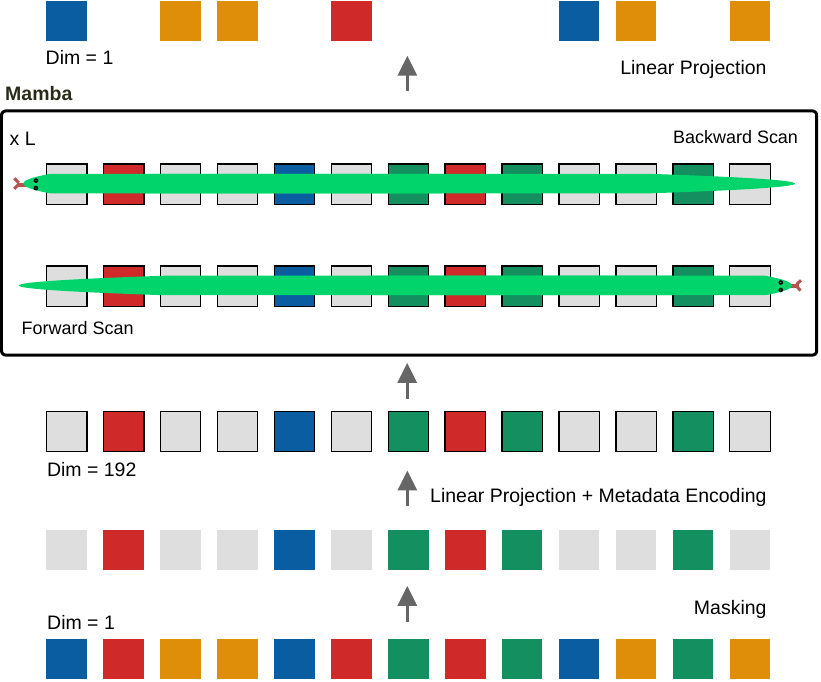}
\caption{Model architecture \includegraphics[height=10pt]{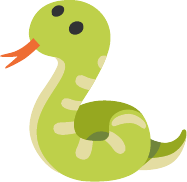}. During training, sparse WSE observations (colored) are sampled and part of these are masked. Learned query tokens (grey) and temporal, satellite, coordinate, and river-tree metadata are added. 
Tokens 
are processed by $N$ bidirectional selective-SSM layers, and
mapped to normalized WSE by satellite-specific heads. In inference, the masking is replaced by the introduction of query tokens carrying the metadata of what is to be predicted.}
\label{fig:architecture}
\end{figure}

\begin{figure}[t]
\centering
\includegraphics[width=0.8\linewidth]{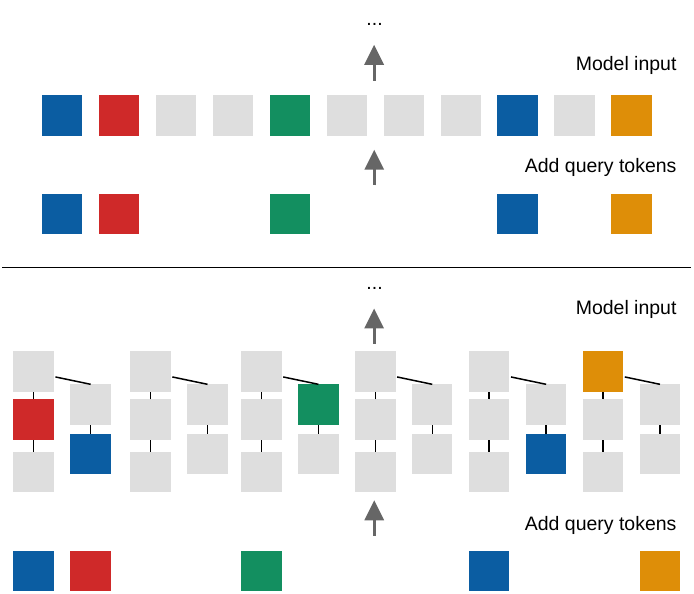}
\caption{Construction of inference sample that is input to our model (top) vs. to graph based methods for imputation like SPIN~\cite{marisca2022spin} or KITS~\cite{xu2025kits}. 
Learned query tokens 
are added to sparse conditioning inputs to inform the model of what to decode.
Graph based methods add large amounts of query tokens (grey) to obtain a full spatiotemporal grid, while we add tokens only for a single location, but still using context from different locations.}
\label{fig:inference-sample}
\end{figure}

\paragraph{Node-Independent Metadata Encodings.}
Each dynamic token (masked and non-masked) receives temporal, source, geographic, and topological metadata:
\begin{align}
\mathbf{p}_\ell
={}&
\mathbf{E}_{\mathrm{month}}[m_\ell]
+\operatorname{PE}_{\mathrm{day}}(t_\ell-t_{\min})
+\mathbf{E}_{\mathrm{src}}[r_\ell] \notag\\
&+\mathbf{W}_{a}\mathbf{c}^{a}_{i_\ell}
+\mathbf{W}_{g}\mathbf{c}^{g}_{i_\ell}
+\operatorname{TreePE}(\mathbf{b}_{i_\ell}).
\label{eq:metadata-encoding}
\end{align}
Here, $m_\ell$ is the month, $\mathbf{E}_{\mathrm{month}}$ and $\mathbf{E}_{\mathrm{src}}$ are month and source embedding tables, $\operatorname{PE}_{\mathrm{day}}$ is a sine/cosine position encoding, $t_\ell-t_{\min}$ is the token's date index offset to the earliest token in the sample, $\mathbf{c}^{a}_{i}$ contains normalized coordinates
relative to the most-downstream sampled reach, $\mathbf{c}^{g}_{i}$ contains absolute
geographic coordinates, and $\mathbf{b}_{i}$ is the local branch path from reach $i$ to
the sampled downstream root. The source embedding distinguishes measurements produced
by different altimeters. 
Metadata are added to model inputs before every layer: $\mathbf{e}_\ell \leftarrow \mathbf{e}_\ell + \mathbf{p}_\ell$.

Each sampled location additionally contributes a static token encoding its mean WSE w.r.t. the most-downstream reach in the sample: $v_i^{\mathrm{mean}}=\operatorname{Norm}_{\beta}(\mu_i-\mu_{i_*})$,
where $i_*$ is the most-downstream sampled reach.
Static metadata tokens form a prefix. Dynamic tokens are then 
sorted so that they are ordered first by day and then along the river topology according to downstream flow rank.

We approximate a local subgraph with a tree and encode the position of reach $i$ by the sequence of branch choices $\mathbf{b}_i=(b_{i,0},\ldots,b_{i,K-1})$ that separate $i$ from the local downstream root following~\citet{shiv2019tree}.
Let $\mathbf{u}_{i,k}\in\{0,1\}^{B}$ be the one-hot
encoding of branch choice $b_{i,k}$, where $B=3$ is the branching factor
and $\mathbf{u}_{i,k}=\mathbf{0}$ for padded path positions.
A decay rate applied to depth $k$ is computed from learned weights $w \in \mathbb{R}^F$, for channel $f=1,\ldots,F$:
\begin{align}
[\mathbf{g}_k]_f
=
\rho_f^k
\sqrt{\frac{F}{2}\left(1-\rho_f^2\right)},
\quad
\rho_f=\tanh(w_f)
\end{align}
The tree encoding is obtained by assigning this weight vector to the branch
selected at each depth:
\begin{equation}
\operatorname{TreePE}(\mathbf{b}_i)
=
\mathbf{W}_{\mathrm{tree}}
\operatorname{vec}
\left(
\operatorname{concat}_{k=0}^{K-1}
\mathbf{u}_{i,k}\mathbf{g}_k^\top
\right).
\label{eq:tree-encoding}
\end{equation}
Each path depth therefore activates one branch-specific block of $F$
components. The learned geometric decay rates allow different channels to
emphasize different parts of the path, while shared path segments receive the
same encoding. 
Learned weights $\mathbf{W}_{\mathrm{tree}}$ project the encoding from $K\cdot B\cdot F$ dimensions to
the model dimension.

These encodings contain no learned lookup indexed by node index (SWORD reach). The same
functions encode coordinates, relative branch paths, source identities, and scalar WSE
metadata at every location. The model is therefore inductive to reaches not observed
during training, provided their SWORD topology and static metadata are available.

\subsection{Bidirectional Mamba \includegraphics[height=10pt]{imgs/model/snake-emoji.pdf}}

The ordered sequence is processed using standard Mamba blocks
\citep{gu2023mamba}. At its core, Mamba applies an input-dependent state-space recurrence
\begin{align}
\mathbf{s}_\ell
&=
\overline{\mathbf{A}}_\ell\mathbf{s}_{\ell-1}
+\overline{\mathbf{B}}_\ell\mathbf{x}_\ell,\\
\mathbf{y}_\ell
&=
\mathbf{C}_\ell\mathbf{s}_\ell+\mathbf{D}\mathbf{x}_\ell,
\label{eq:mamba-core}
\end{align}
where the discretization and the maps
$\overline{\mathbf{B}}_\ell$ and $\mathbf{C}_\ell$ depend on the current input
$\mathbf{x}_\ell$. This selectivity allows the model to retain or discard information as
it scans the sequence. 

Each residual block has the form
\begin{equation}
\operatorname{MBlock}(\mathbf{X})
=
\mathbf{X}
+
\operatorname{Mamba}\!\left(\operatorname{LN}(\mathbf{X})\right).
\end{equation}
The standard formulation of Mamba is 1-directional, but we combine a forward and backward scan, as proposed by~\citet{zhu2024vision}, but in our case to integrate past information from upstream nodes with future information from downstream nodes.
After restoring the reverse output to the original order, both directions
are fused:
\begin{equation}
\mathbf{H}^{(k)}
=
\mathbf{W}_{\mathrm{bi}}
\left[
\mathbf{H}^{(k)}_{\rightarrow}
\,\Vert\,
\operatorname{rev}_{\mathrm{dyn}}
\left(\mathbf{H}^{(k)}_{\leftarrow}\right)
\right].
\label{eq:bidirectional-mamba}
\end{equation}
Only the dynamic sequence is reversed; the static metadata prefix remains fixed.
The resulting model uses observations on both sides of a query and is therefore a
non-causal reconstruction model rather than a forecaster. A source-specific linear head
maps each final token representation to normalized WSE prediction $\widehat z_\ell$.

\subsection{Training}

Training combines location masking, which hides complete source-location time series to teach spatial reconstruction at unseen locations, and
random masking, which hides individual observations to teach temporal densification. Static metadata remain visible. If
$\Omega$ is the set of masked, non-padding observations in a minibatch, we minimize
\begin{equation}
\mathcal{L}(\theta)
=
\frac{1}{|\Omega|}
\sum_{\ell\in\Omega}
\left(\widehat z_\ell-z_\ell\right)^2.
\label{eq:masked-mse}
\end{equation}

\subsection{Inference}
\label{sec:inference}

To predict reach $i$ over interval $I$, we use it as anchor $a = i$ to construct a subgraph sample as described above, and we add $i$ as virtual location containing exactly one
masked query token $(i,t)$ for each day $t\in I$. No daily queries are introduced for the
other reaches in the sampled neighborhood: their sparse satellite measurements remain
the observed context,
\begin{equation}
\mathcal{D}^{\mathrm{inf}}_{i,I}
=
\mathcal{D}_{i,I}
\cup
\left\{(i,t,\textrm{masked}):t\in I\right\}.  %
\label{eq:inference-sample}
\end{equation}
Predicting only the target reach prevents the sequence from being dominated by virtual
tokens and maintains a favorable ratio of observed context to queries even under extreme
graph sparsity (illustrated in Figure~\ref{fig:inference-sample}).
Each window is reconstructed independently, without feeding
predictions back into the model, and we average overlapping windows.

\section{Experiments}

\begin{table}[t]
\centering
\small
\begin{tabular}{lrcc}
\toprule
& &  \multicolumn{2}{c}{\textbf{RMSE\,$\downarrow$}} \\
\textbf{Model} & \textbf{} & {>2023/07} & {<2022/06} \\ %
\midrule
\multicolumn{4}{l}{\emph{Transductive}}  \\
\multicolumn{1}{l}{Temp. LSTM} & \textit{sub}  & 1.47  & 2.47  \\
GRIN  & \textit{full} & 0.88  & 1.28  \\
& \textit{sub} & 2.52  & 2.25  \\
SPIN-H  & \textit{full} & 0.93  & 1.58  \\
& \textit{sub} & 0.74  & 0.94  \\
ImputeFormer  & \textit{full} & 1.69  & 1.56  \\
& \textit{sub} & 0.67  & 0.91 \\
\midrule
\multicolumn{4}{l}{\emph{Inductive}} \\
kNN & \textit{full}   &  1.77 & 2.24 \\
IGNNK & \textit{full} & 2.33 & 2.38   \\
& \textit{sub} & 1.11  & 1.34  \\
KITS & \textit{full} & 2.38  & 2.40  \\
& \textit{sub} & 1.23  & 1.61  \\
\midrule
\textit{Ours} \includegraphics[height=10pt]{imgs/model/snake-emoji.pdf} & \textit{sub} & \textbf{0.62}  & \textbf{0.84}  \\
\bottomrule
\end{tabular}
\caption{Baseline comparison on held-out in situ gauges for the 2.7\% (>2023/07) and 0.6\% (<2022/06) sparsity regimes. \textit{sub}: subgraph sampling and \textit{full}: full graph in each sample.}
\label{tab:architectures}
\end{table}

\begin{figure}[t]
\centering
\includegraphics[width=0.75\linewidth]{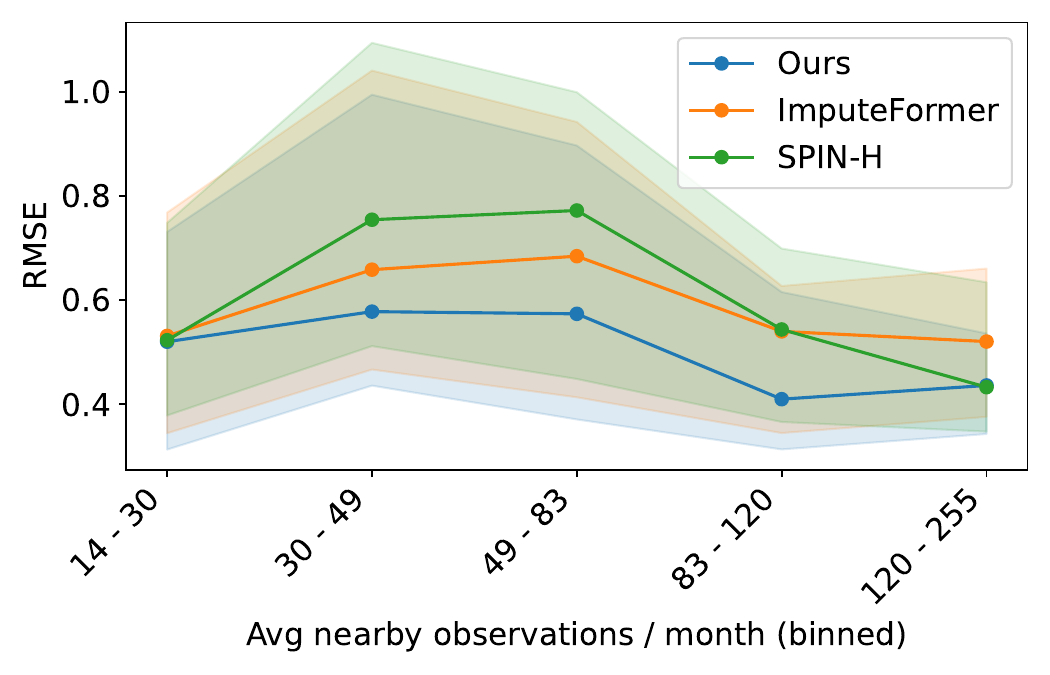}
\caption{{Ours} vs. baseline RMSE as a function of the number of observations in the subgraph neighborhood per month.}
\label{fig:metrics_vs_obs}
\end{figure}

Hyperparameters are listed in Appendix~\ref{app:implementation}. 
Our model has 3.3M parameters and takes 2~h to train with less than 4 GB of GPU memory on one H100,
predicting all 19K nodes takes another 1,5~h.
Training data spans 2023-07--2026-05 and combines SWOT with HydroWeb and ICESat-2 observations. 
For our training and for baselines, time series with less than 10 (for ICESat-2) or 20 (for the remaining sources) values are discarded, reducing the number of observed ICESat-2 locations from 18.5K to 10K (the data are kept in the published dataset for subsequent works to use as they see fit). With $\sim$75\% of reaches observed by the remaining time series, our task involves both transduction and induction, even though 95\% of the evaluation gauges are located on a reach that is observed (this can be expected to slightly disadvantage transductive methods).
We evaluate in two temporal settings:
\begin{enumerate}
    \item \textbf{SWOT period} (2023-07 to 2026-05): 
    The sparsity of this period is lower due to the presence of SWOT measurements (in both training and inference): 2.7\% observed.
    \item \textbf{Hindcast} (2016-01 to 2022-06-30): the model must extrapolate backward in time using only classical altimetry from HydroWeb and ICESat-2 observing 0.6\% of day--reach slots, without any SWOT data.
\end{enumerate}
We keep gauge data from the year 2022-07 to 2023-07 as evaluation setting for ablations (not for early stopping).
The model never sees in situ data during training, preventing leakage also when the training and inference periods coincide.

\paragraph{Baselines.}
We select baselines to cover complementary settings and model families. A
per-reach bidirectional LSTM inspired by BRITS-I~\citep{cao2018brits} and spatial kNN isolate
temporal modeling and non-parametric spatiotemporal interpolation. 
Among transductive
imputers, GRIN~\citep{cini2022grin} represents recurrent graph message passing, while
SPIN-H~\citep{marisca2022spin} and ImputeFormer~\citep{nie2024imputeformer} represent
attention-based approaches and demonstrated high (95\%) point sparsity. 
IGNNK~\citep{wu2021ignnk} and KITS~\citep{xu2025kits} test inductive reconstruction at unseen reaches; KITS is especially
relevant because it addresses the gap between sparse training graphs and virtual targets at
inference. 
Some baselines natively support multivariate inputs; for those that do not, we add
separate source channels, compare them with merging observations into one WSE series, and report
whichever performs best.
We run graph baselines both on the
full graph and on the same sampled subgraphs as our model.
We also compare with Reach-Reg~\citep{halicki2026reachreg}, the domain-specific
baseline for SWOT WSE densification,
which we fit on SWOT RiverSP data to make predictions for our gauges.
Appendix~\ref{app:baselines} describes baseline details.

\section{Results}

\paragraph{Baseline Comparison.}
Table~\ref{tab:architectures} compares our model with existing ML methods; we outperform all methods in all settings.
It also shows that baselines have a larger gap between performance on SWOT period imputation where 2.7\% of days have an input measurement and on the Hindcast setting where only 0.6\% is observed.
SPIN-H and ImputeFormer with subgraphs reach good accuracy, which is in line with their reported improved results on sparse settings compared to other models.
Except for GRIN, subgraph sampling outperforms handling the full graph at once, we attribute this to our larger graph compared to those of existing datasets.
The transductive baselines do better than inductive baselines (but not than our model which is also inductive).
Figure~\ref{fig:metrics_vs_obs} shows performance of our model vs. ImputeFormer and SPIN-H, in function of number of the average amount of context tokens available in the subgraph (colored tokens in Figure~\ref{fig:inference-sample}), where baseline performance deteriorates faster when less conditioning observations are available.
These results support the hypothesis that sequence models outperform GNN-based approaches under extreme spatiotemporal sparsity.

\begin{table}[t]
\centering
\small
\begin{tabular}{llcc}
\toprule
\textbf{Configuration} & & \textbf{RMSE\,$\downarrow$}  \\
\midrule
\emph{Base}  & & \textbf{0.56}   \\
\midrule
\emph{Metadata enc} & No tree encoding          & 0.58  \\
& No satellite enc         & 0.58   \\
& No mean WSE tokens        & 0.58   \\
\midrule
\emph{Data sources} & No ICESat-2       & 0.59   \\
& No SWOT \& ICEsat-2  & 0.60   \\
\midrule
\emph{Token order} 
& Flow                & 0.57   \\
& Random              & 0.64  \\
\midrule
\emph{Architecture} & 1-directional & 0.72  \\
\quad \emph{\& inputs} & Isolated & 2.58  \\
\bottomrule
\end{tabular}
\caption{Ablation study (validation period, mean over held-out gauges). \textit{Base} has all metadata encodings and tokens, uses all 3 data sources in training and inference (when available), sees input sequences sorted by time, and is 2-directional.}
\label{tab:ablations}
\end{table}

\begin{figure}[t]
\centering
\includegraphics[width=0.75\linewidth]{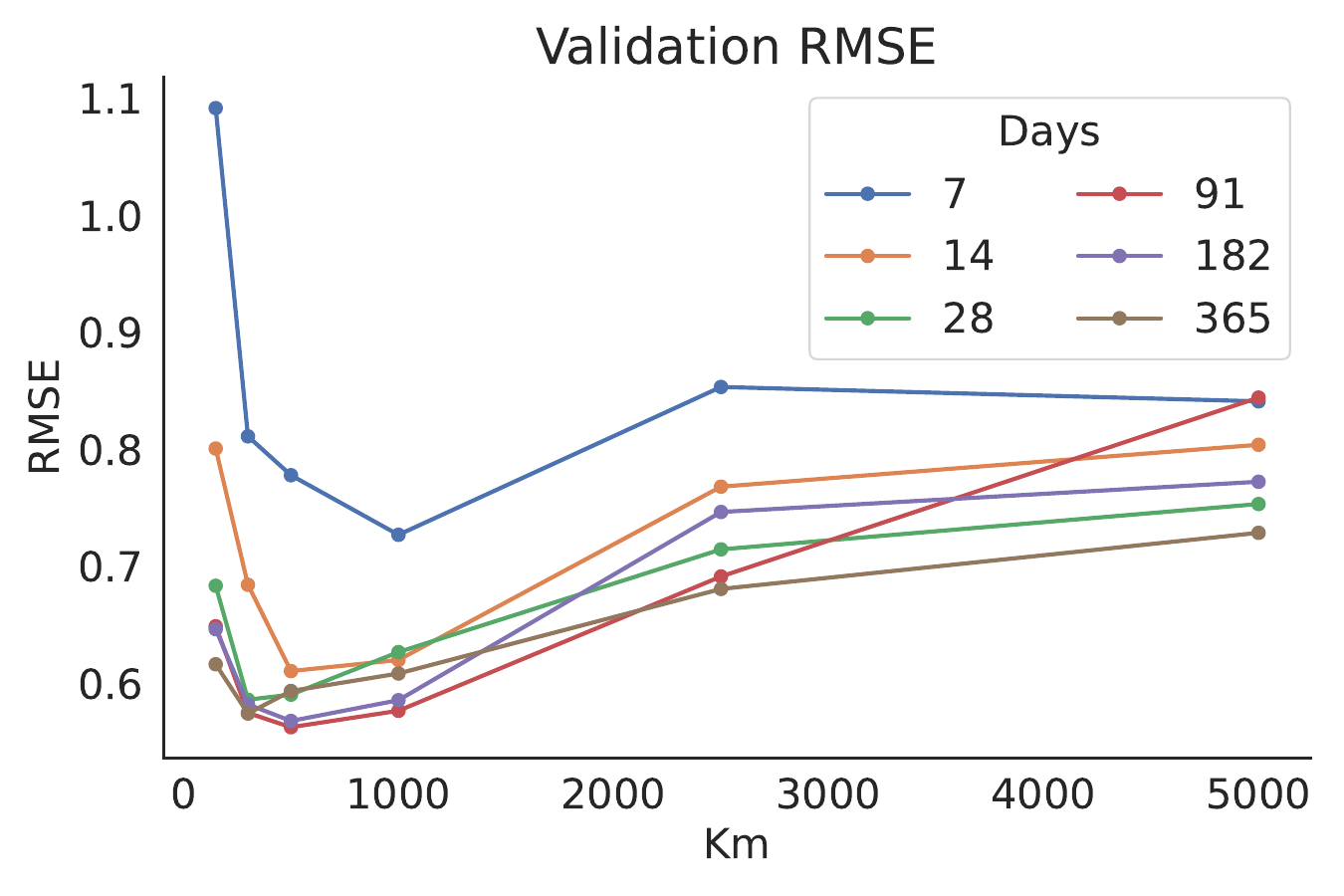}
\caption{Ablation of sampled subgraph size limits.}
\label{fig:neighborhood-size}
\end{figure}

\paragraph{Ablations.}
Table~\ref{tab:ablations} presents ablation experiments on metadata encoding and data source inclusion.
In contrast to~\citet{kirschstein2024river}, we find that graph topology and spatial context do help: the metadata encodings as well as tokens being ordered by time or flow improve predictions slightly and the subgraph model (\textit{Base)} significantly outperforms the \textit{Isolated} model (that sees only 1 node timeseries).
All data sources (SWOT, ICESat-2 and HydroWeb) contribute to better performance.
Figure~\ref{fig:neighborhood-size} shows a comparison of time window size and the spatial neighborhood size that the subgraph is sampled from: a neighborhood size of 300-1000~km works best and performance is robust to temporal window size, with 3~months giving the lowest RMSE.

\begin{table}[t]
\centering
\small
\setlength{\tabcolsep}{1mm}
\begin{tabular}{lrcccccc}
\toprule
& & \multicolumn{3}{c}{\textbf{RMSE}\,$\downarrow$}
& \multicolumn{3}{c}{\textbf{KGE}\,$\uparrow$} \\
\cmidrule(lr){3-5} \cmidrule(lr){6-8}
\textbf{Period} & \textbf{Cov.} & RR & IF & \includegraphics[height=10pt]{imgs/model/snake-emoji.pdf} & RR & IF & \includegraphics[height=10pt]{imgs/model/snake-emoji.pdf} \\
\midrule
SWOT era
  & 184/284
  & 0.90 & 0.61 & \textbf{0.55}
  & 0.85 & 0.92 & \textbf{0.94} \\
Hindcast
  & 149/250
  & 0.85 & 0.80 & \textbf{0.70}
  & 0.88 & 0.88 & \textbf{0.91} \\
\bottomrule
\end{tabular}
\caption{Comparison with Reach-Reg (RR) and ImputeFormer (IF). Mean metrics are
computed on gauges overlapping with Reach-Reg; \textbf{Cov.} reports
RR/total gauge coverage. Scores differ from other tables because we evaluate only on gauges for which Reach-Reg makes a prediction.}
\label{tab:results_reachreg}
\end{table}

\paragraph{Comparison with Reach-Reg.}
Table~\ref{tab:results_reachreg} compares our model against Reach-Reg and ImputeFormer on overlapping in situ gauges across both evaluation periods.
Kling--Gupta Efficiency~\citep{gupta2009kge} is defined as
\(\text{KGE}=1-\sqrt{(r-1)^2+(\alpha-1)^2+(\beta-1)^2}\), where \(r\) is the
Pearson correlation between predictions and observations,
\(\alpha=\sigma_{\mathrm{pred}}/\sigma_{\mathrm{obs}}\) measures variability, and
\(\beta=\mu_{\mathrm{pred}}/\mu_{\mathrm{obs}}\) measures bias. It is widely used in Hydrology and jointly evaluates
correlation, variability, and mean agreement, with a perfect score of 1 and higher values indicating
better performance.
Our model outperforms both methods on both periods and provides greater coverage than Reach-Reg.
We evaluate Reach-Reg for making hindcasts, even though it was originally developed only to make predictions for the SWOT-era (details in Appendix~\ref{app:baselines}).
As a state-of-the-art hydrology method, 
it sets a level of accuracy along with a coverage baseline that new methods should aim to meet or surpass on the \texttt{AmazonWSE} benchmark.
Visualizations and comparisons of predictions are provided in Appendix~\ref{app:predictions}.

\section{Conclusion}

We introduced \texttt{AmazonWSE}, a benchmark dataset for spatiotemporal graph imputation on time series of water levels observed satellite altimetry, covering the Amazon river basin, with 19K river reaches as nodes, 99\% sparsity, and a directed acyclic graph topology.
We showed that prior spatiotemporal graph imputation methods are not adapted to this scale and sparsity, and proposed a bidirectional selective state space model that outperforms them by sampling connected subgraphs and flattening space and time into a single token sequence.
Our model also outperforms a state-of-the-art non-neural baseline for SWOT-based WSE densification and produces denser reconstructions.
We hope that these results inspire the community to develop improved methods for spatiotemporal imputation in extremely sparse settings such as river monitoring from satellite altimetry.

\bibliography{aaai2027}

\clearpage

\appendix
\setcounter{secnumdepth}{2}
\section{Appendix}

\paragraph{Overview of Supplementary Material}\begin{itemize}
    \item This Appendix (\textit{Technical Supplement}):\begin{enumerate}
        \item Dataset Sources
        \item Data Preprocessing Details
        \item Storage Format
        \item Hyperparameters and Model Details
        \item Baselines
        \item Predictions
    \end{enumerate}
    \item \textit{Media Supplement}: animated .gif files of 1) sparse ICESat-2 and HydroWeb inputs and 2) imputation prediction of daily WSE for all SWORD reaches by our model.
    \item \textit{Code and Data Supplement}: \begin{itemize}
        \item Netcdf files for each source in the \texttt{AmazonWSE} dataset with a tiny amount of measurements: as an example of the storage format.
        \item Code of data export and preprocessing scripts.
        \item Code to run training and evaluation of our model and the baselines on the tiny included dataset.
    \end{itemize}
\end{itemize}

\subsection{Dataset Sources}
\label{app:statistics}

\begin{figure*}[t]
\centering
\includegraphics[width=0.499\linewidth]{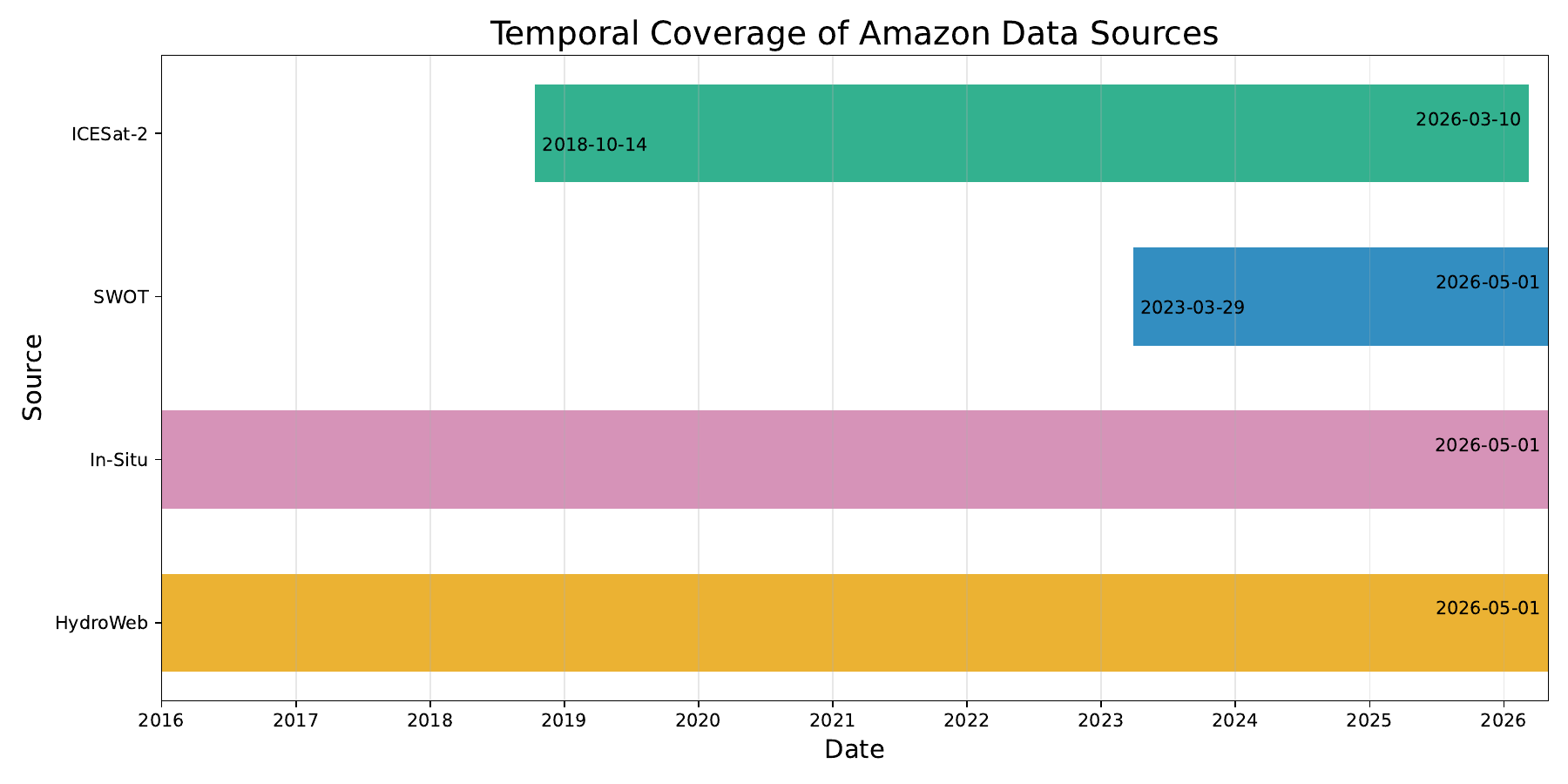}\hfill
\includegraphics[width=0.499\linewidth]{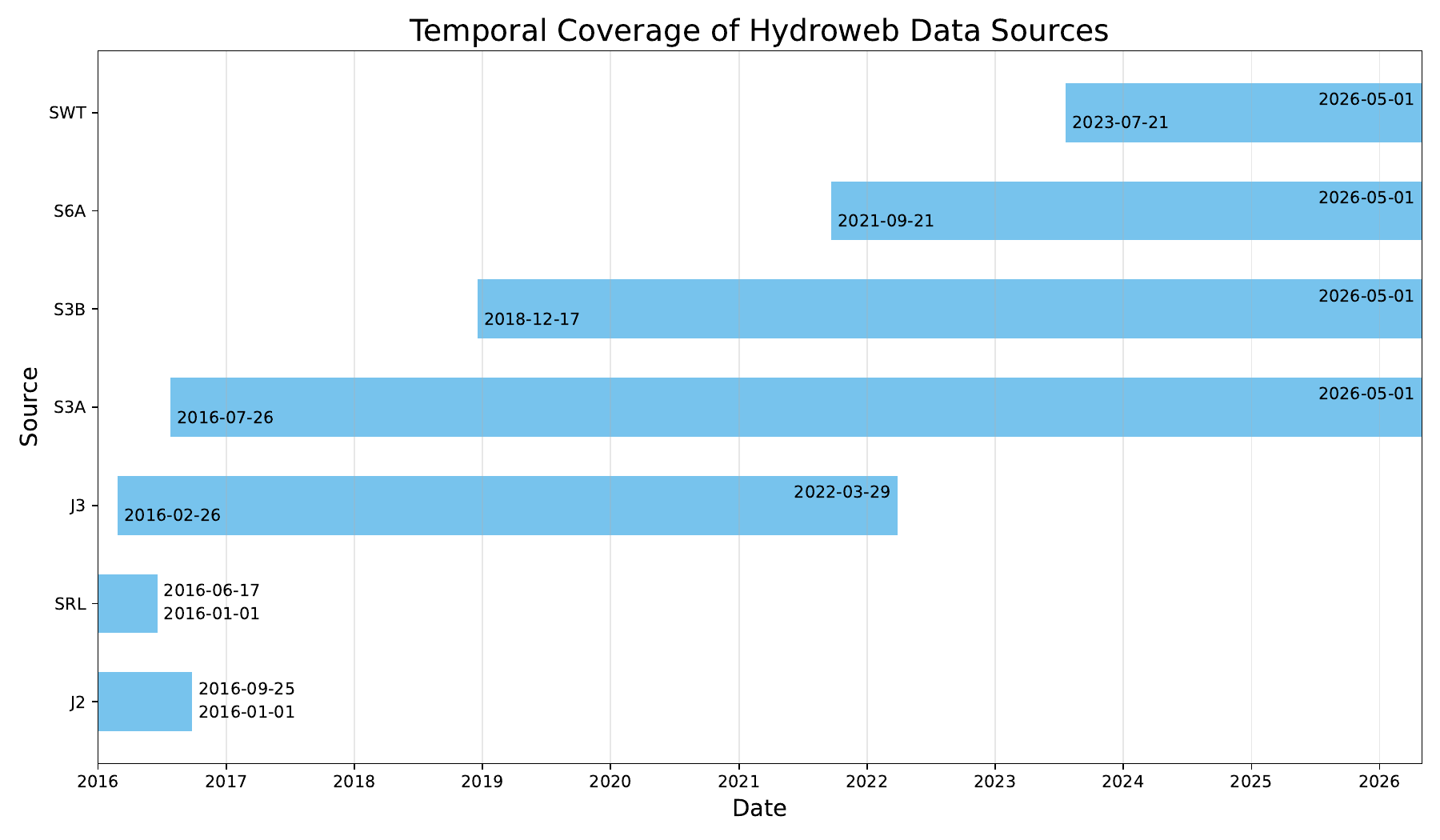}
\caption{Left: temporal coverage of sources in \texttt{AmazonWSE}. Right: temporal coverage of satellite sensors included in HydroWeb: the SWOT nadir altimeter (i.e., a different sensor than the swath altimetry sensor used for the SWOT data source in \texttt{AmazonWSE}), Sentinel-3A/B, Sentinel-6A, JASON-2/3, SARAL.}
\label{fig:temporal-coverage-sources}
\end{figure*}

Figure~\ref{fig:temporal-coverage-sources} shows the temporal coverage of different sources in \texttt{AmazonWSE} and in HydroWeb.
The next paragraphs provide more details about the sources when relevant.

\paragraph{SWORD.}
The Surface Water and Ocean Topography (SWOT) Mission River Database (v17b) was designed as a complementary static dataset to SWOT products and provides a topological graph of 19,172 river reaches in the Amazon basin \cite{altenau2021sword}.\footnote{\url{https://zenodo.org/records/15299138}.}

\paragraph{ICESat-2.}
The NASA Ice, Cloud and land Elevation Satellite-2 (ICESat-2) mission carries the  ATLAS (Advance Topographic Laser Altimeter System) instrument. 
Data is collected along three pairs of narrow laser beams that operate at 532nm wavelength. 
The beams in the pair are separated by a distance of around 90m, while each pair is separated by the next by a distance of 3 km. 
The data in \texttt{AmazonWSE} are from the Level3B ATL22 Mean Inland Surface Water Data product version 4~\cite{jasinski2025atl22}, which reports mean along-track surface elevation for each beam for each transect across a water body. 
Our observations span Oct. 2018--Mar. 2026 because the product has been released up until March 2026 at the time of writing.

\paragraph{SWOT RiverSP.}
We use data from the ``SWOT Level 2 River Single-Pass Vector Reach Data Product, Version D'', and of the science cycle that started in July, 2023~\cite{swot}.

\paragraph{HydroWeb.next.}
We use the operational and research collections~\citep{cretaux2015hydroweb} and include data from the following satellites: the SWOT nadir altimeter (i.e., a different sensor on the same satellite than the InSAR swath altimetry sensor used for the SWOT data source in \texttt{AmazonWSE}), Sentinel-3A/B, Sentinel-6A, JASON-2/3, SARAL.

\subsection{Data Preprocessing Details}
\label{app:preprocessing}

\begin{figure*}[t]
\centering
\includegraphics[width=0.49\linewidth]{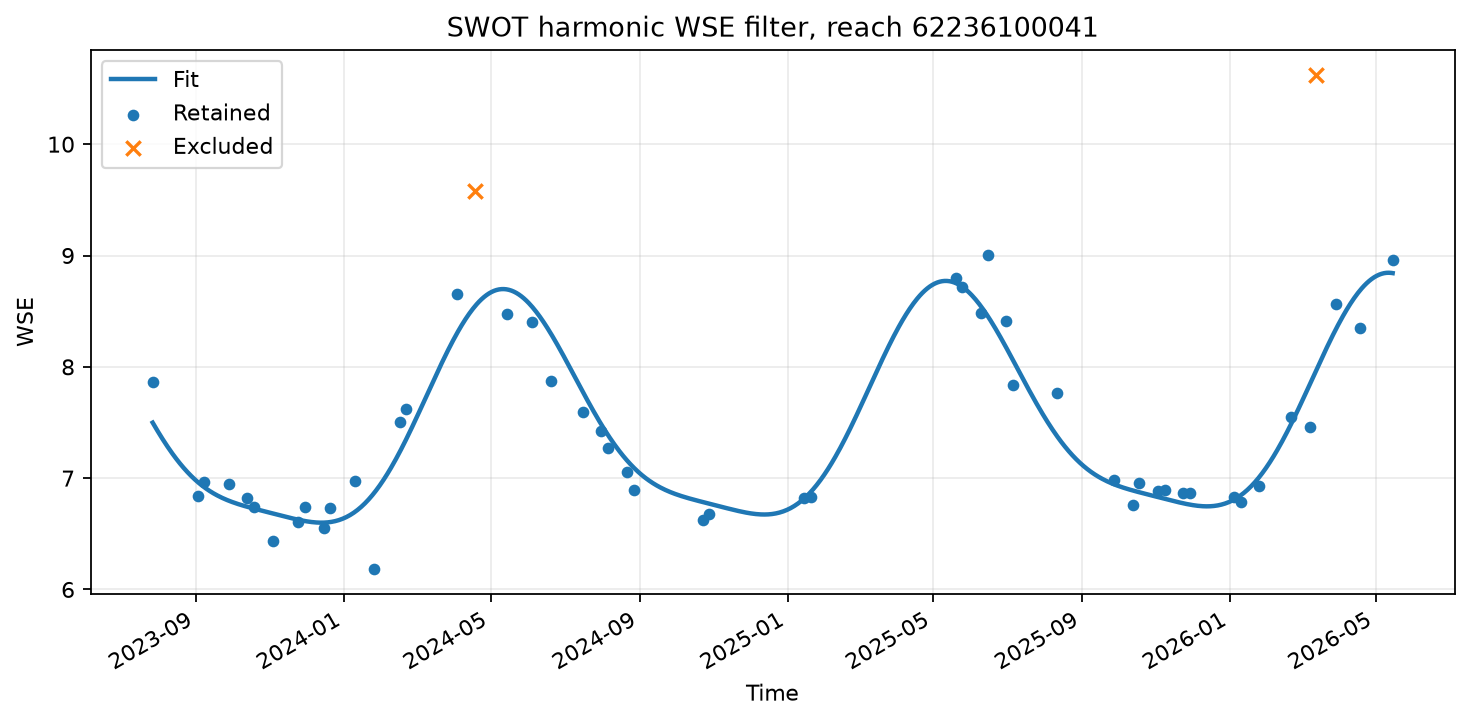}\hfill
\includegraphics[width=0.49\linewidth]{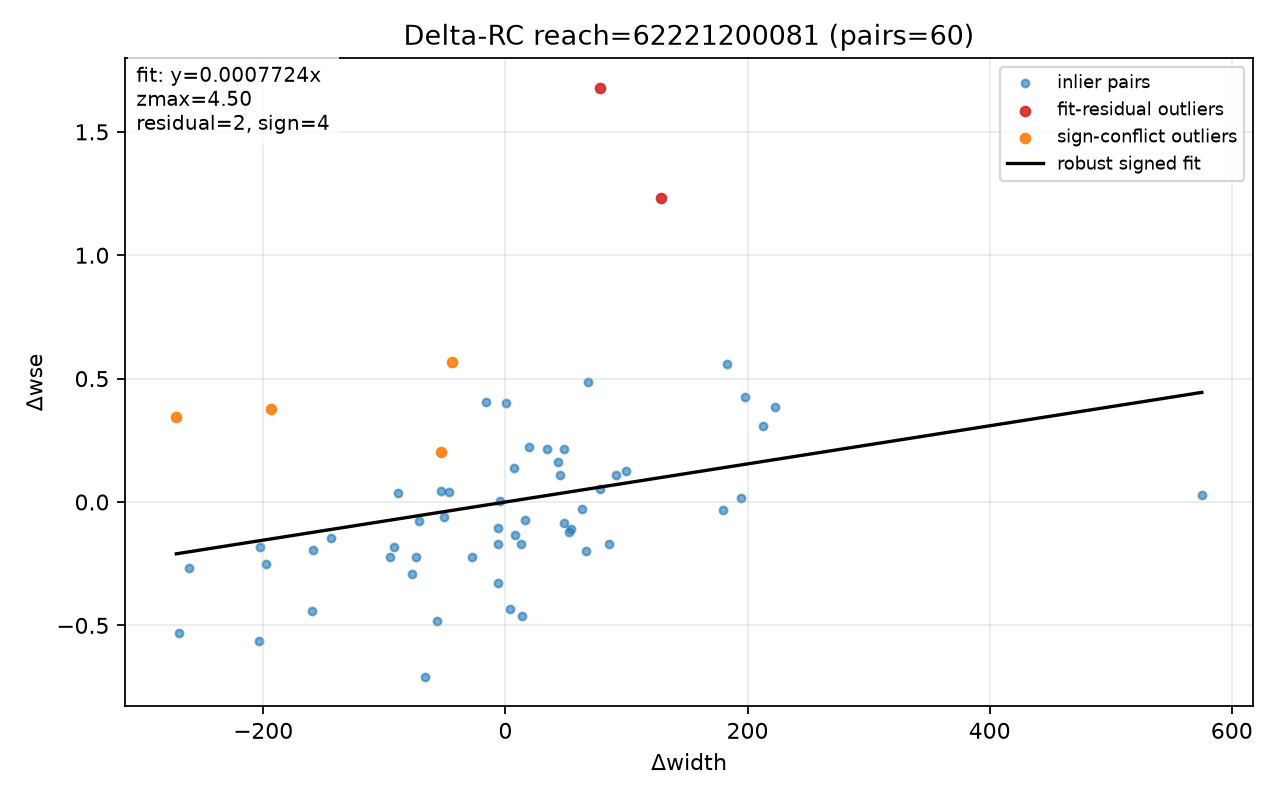}
\caption{
Examples of SWOT observations retained and rejected by the harmonic-residual (left). 
Observations retained and discarded by the rating curve filter that considers $\Delta\text{width}/\Delta\text{WSE}$ (right).
}
\label{fig:swot-filters}
\end{figure*}

\begin{figure*}[t]
\centering
\includegraphics[width=0.499\linewidth]{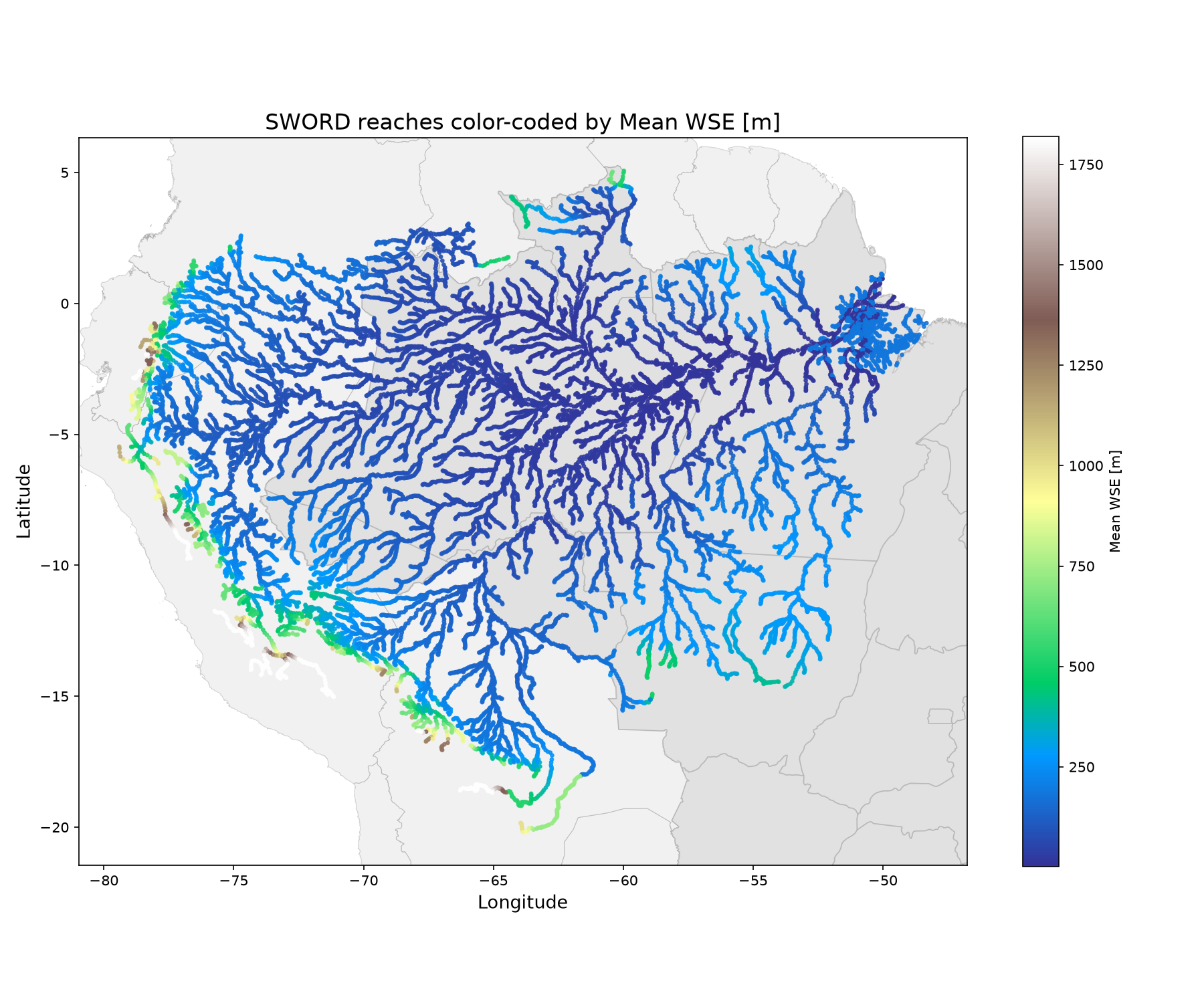}\hfill
\includegraphics[width=0.499\linewidth]{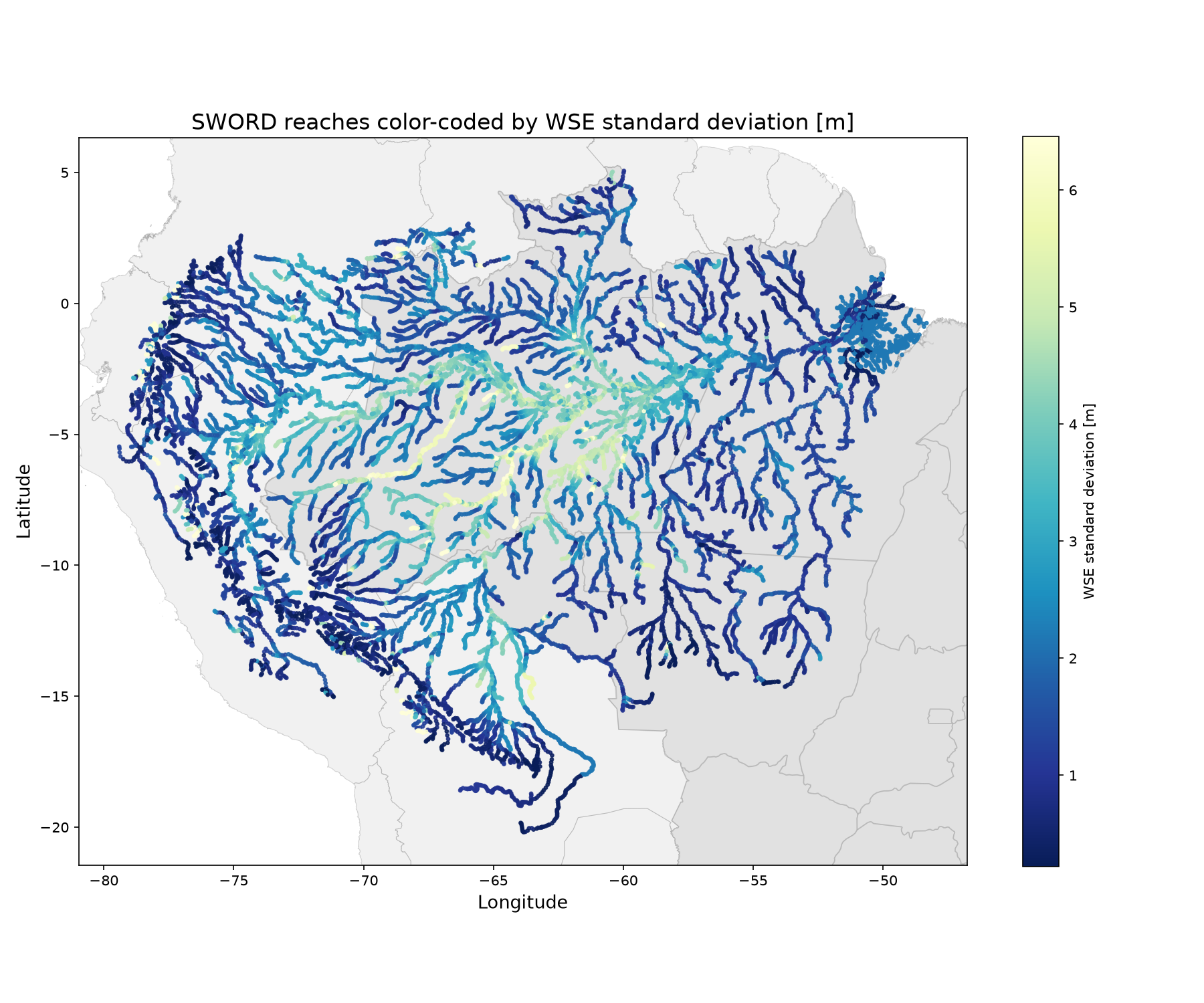}
\caption{
Interpolated SWOT mean and std fields that are used to de-normalize predictions.
}
\label{fig:mean-fields}
\end{figure*}

\begin{figure*}[t]
\centering
\includegraphics[width=0.499\linewidth]{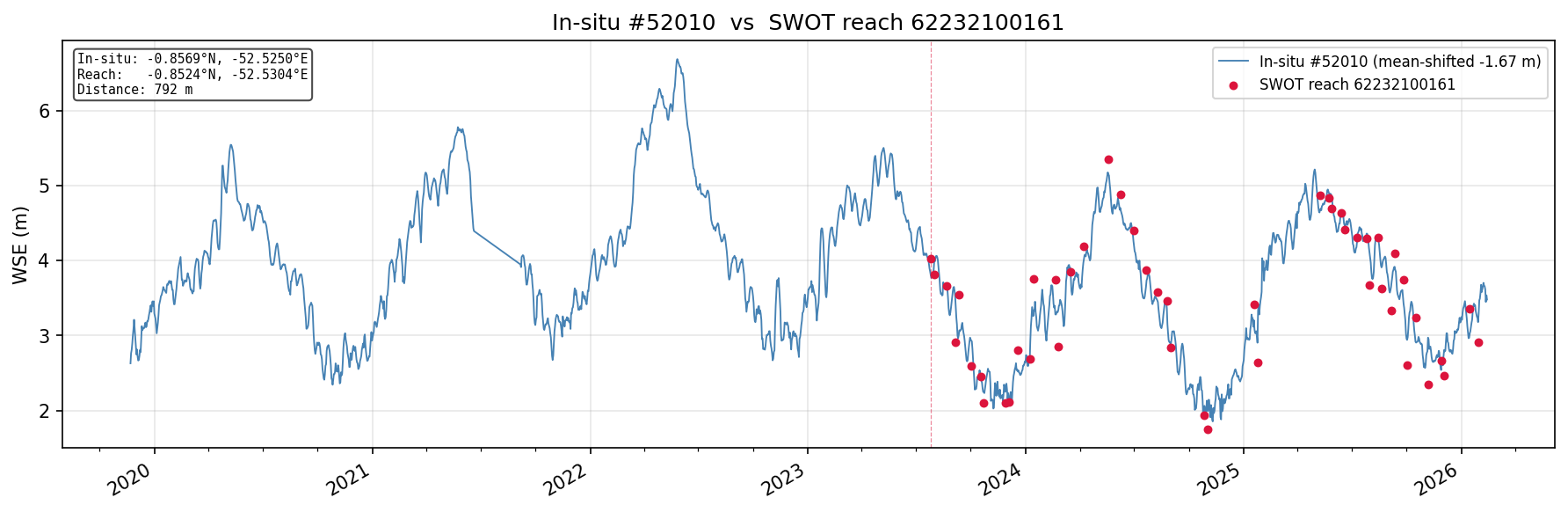}\hfill
\includegraphics[width=0.499\linewidth]{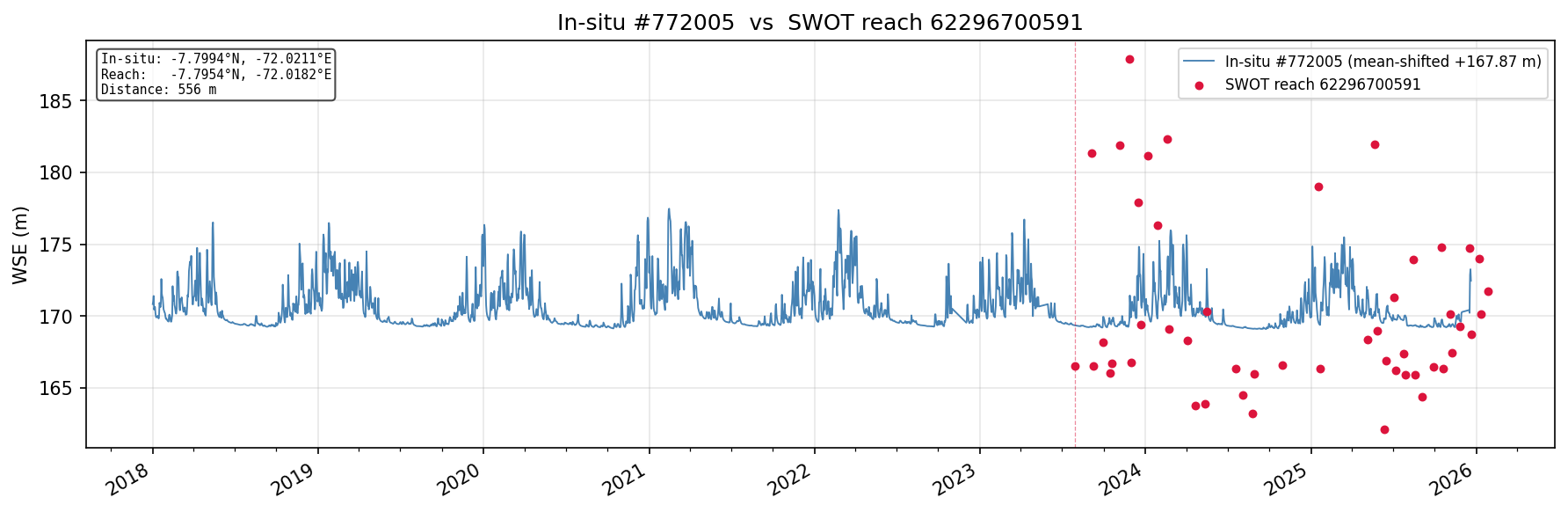}
\includegraphics[width=0.499\linewidth]{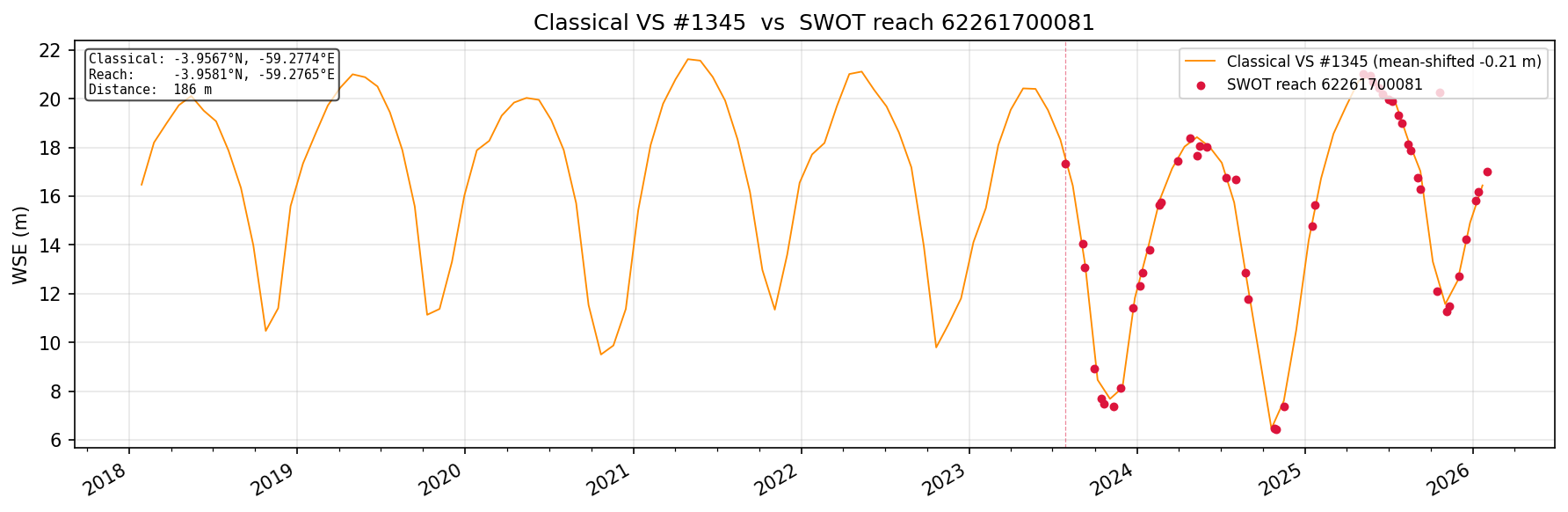}\hfill
\includegraphics[width=0.499\linewidth]{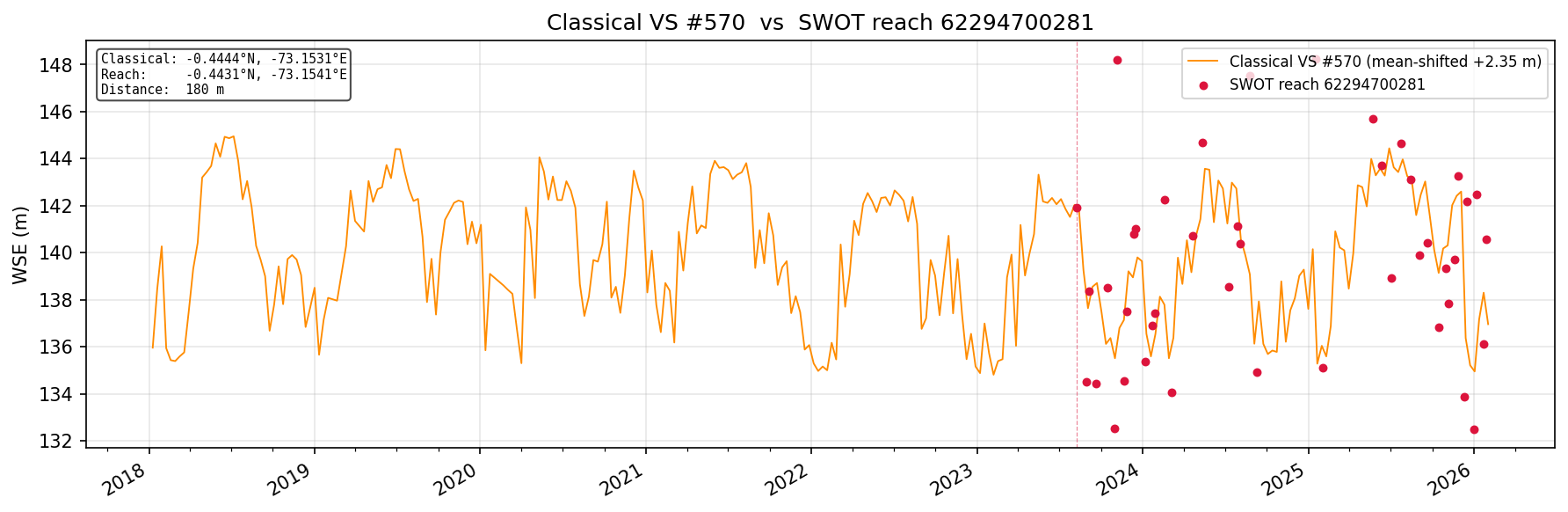}
\caption{
Unfiltered SWOT RiverSP measurements (in red) with time series of collocated in situ gauges (top) and HydroWeb locations (bottom).
While for some reaches SWOT measurements follow HydroWeb/in situ gauges closely (left), for other reaches, SWOT measurements are very noisy or low quality (right).
}
\label{fig:swot-comparisons}
\end{figure*}

This section describes the data processing in more detail.
All timestamps are first converted from local time to UTC (if not already in UTC).
Subsequently, all measurements are assigned to daily bins on the shared temporal grid.

A quality flag in the shared data files records missing, rejected (by quality filters), and retained observations.

\subsubsection{SWOT.}

Our product-level screening draws on the criteria used by \citet{andreadis2025swot} and \citet{halicki2026reachreg}, with additional criteria added by ourselves.
RiverSP records 
are aggregated into 24-hour bins. WSE and width use the median within each
    reach--day, provided measurement standard deviations are aggregated accordingly, and quality bit fields use the maximum so that severe flags are
    preserved.
Many filters use variables shared as part of the SWOT RiverSP products, the names of those variables are marked here by the monospaced font, like \texttt{wse\_r\_u}.

Altogether filters take away about 50\% of measurements.
This may seem like a lot, but examples shown in Figure~\ref{fig:swot-comparisons} of unfiltered RiverSP SWOT time series along with collocated HydroWeb (``classical'' altimetry) and in situ gauge time series can hopefully motivate the need for this extensive filtering.

\paragraph{Filters that Apply to Reaches.}
\begin{itemize}
    \item \textbf{River Width.} We remove reaches with a static (mean) width in SWORD lower than 50~m. \citet{andreadis2025swot} used an 80~m cutoff for their early
    discharge validation; we instead use SWOT's 50~m river-observation goal
    \citep{biancamaria2016swot,andreadis2025swot} to retain the mission's intended
    lower-width range.

    \item \textbf{Reach Identifier.} 
    SWORD assigns identifiers to each reach of the form ``CBBBBBRRRRT'', where the ending ``T'' encodes a reach type. The entire identifier follows the Pfafstetter coding system where the characters encode the topology of the river network~\citeapp{verdin1999topological}.
    We remove reaches with $T\in \{4,5,6\}$, since they encode ``dam or waterfall'', ``unreliable topology'', or ``ghost reach/node''.
\end{itemize}
\paragraph{Filters that Apply to Individual Measurements.}
\begin{itemize}
    \item \textbf{Cross-track Distance.} Observations within
    15~km cross-track distance (\texttt{x\_trk\_dist}) of nadir ground track are removed, following~\citet{andreadis2025swot}.
    
    \item \textbf{Crossover Calibration and WSE Uncertainty.}
    Observations with crossover-calibration quality flag \texttt{xovr\_cal\_q} $>$ 1 or with the random component of their reported WSE uncertainty \texttt{wse\_r\_u} $>$ 0.5~m are removed, following~\citet{andreadis2025swot}.
    The \texttt{wse\_u} field provided in the RiverSP products (which estimates the standard deviation of the WSE measurement error of the SWOT sensor) is stored in our dataset's netcdf files (after root-sum-square aggregation if a reach is observed more than once on a single day).

    \item \textbf{Reach Quality Flags.} Observations with
    reach quality flag \texttt{reach\_q} above 2 are removed, following~\citet{halicki2026reachreg}.
    Unlike ~\citet{halicki2026reachreg}, who remove bitwise quality flags \texttt{reach\_q\_b} $>$ 2097152, we remove only \texttt{reach\_q\_b} $>$ 8388608 which additionally allows observations flagged as coming from a lake, but still removes severe flags like outliers.

    \item \textbf{Observed Precision.} 
    Instead of applying a fixed threshold on the fraction of dark pixels (pixels for which no or not enough backscattered signal was measured and for which hence no measurement could be derived that is fed into the RiverSP measurement calculations) like~\cite{andreadis2025swot,halicki2026reachreg}, we estimate whether the observed water pixels support a target precision $p=0.1$~m. 
    We approximate observed-pixel errors to be unbiased, independent, and approximately equal-variance, and normally distributed.
    If $N^{\mathrm{obs}}_{i,t}$ is the number of observed water pixels and
    $u_{i,t}$ is reported WSE uncertainty, an observation is retained only if:
    \begin{align*}
        2\frac{u_{i,t}}{\sqrt{N^{\mathrm{obs}}_{i,t}}} &\leq p \\
        N^{\mathrm{obs}}_{i,t} &\geq
        4\left(\frac{u_{i,t}}{p}\right)^2,\\
        p&=0.1\ \mathrm{m}.
    \end{align*}
    Thus a larger reported uncertainty requires support from more observed pixels, where the factor of 2 approximates a two-sided 95\% normal confidence interval.
    We crudely estimate $N^{\mathrm{obs}}_{i,t}$ using the resolution of the InSAR sensor ($\sim$7.5~m $\times$ 20~\text{m} and the static mean river width and reach length reported in SWORD (thereby assuming that the entire reach falls within one of the two 50~km wide observed swaths by the sensor).

    \item \textbf{Deviation from Harmonics.} 
    For reaches with sufficient
    observations, a two-harmonic seasonal model is fitted and WSE residuals beyond $4.5$
    standard deviations from the mean (residual) are removed (Figure~\ref{fig:swot-filters}, left).
    
    \item \textbf{Rating Curve.} A robust delta rating-curve filter additionally
    identifies inconsistent changes in river width and WSE as observed by SWOT (Figure~\ref{fig:swot-filters}, right; river width and the water level are bound by a monotonously increasing relationship determined physically by the shape of the river bed). 
    We fit a regression line on pairs of $\Delta\text{width}, \Delta\text{WSE}$ between subsequent observations of the same reach, then remove observations of which the $\Delta\text{width}, \Delta\text{WSE}$ deviates more than 4.5 standard deviations from the regressed line, or observations of which the $\Delta\text{width}$ and $\Delta\text{WSE}$ have non-matching signs (allowing a noise buffer).

    \item \textbf{Pass Correction.} We investigate the bias between overpasses
    with different IDs that observe the same reach and found that the bias between passes is minimal.
    We try to correct it by shifting the mean WSE
    observed by the same pass ID towards each other, but found that that it makes negligible difference. 

    \item \textbf{Mean/Std Field Interpolation.}
    Because SWOT provides the densest coverage, we compute its per-location mean and standard deviation, and interpolate this linearly and spatially to obtain dense fields as shown in Figure~\ref{fig:mean-fields}. The interpolation uses inverse distance weighting and averages upstream branch contributions if there are several. We use these fields for de-normalization of predictions (since predictions may have to be made for locations where no measurements are available).
    
\end{itemize}

\subsubsection{HydroWeb.}

The HydroWeb database already cleans and homogenizes measurements as described in~\cite{santosdasilva2010water,normandin2018altimetry}.

\begin{itemize}
    \item Text files returned from the HydroWeb.next API\footnote{\url{https://github.com/CNES/py-hydroweb}} (for the operational and research collections, or \texttt{HYDROWEB\_RIVERS\_OPE} and \texttt{HYDROWEB\_RIVERS\_RESEARCH}) with time series are parsed, heights encoded as missing or fill values are converted to
    consistent missing data indicators; otherwise, HydroWeb's supplied measurements and uncertainties are
    preserved without an additional value-level outlier filter.
    \item Jason-2 Interleaved (J2N) is excluded because its Oct. 2016--May 2017 record is
    too short for stable statistics. SARAL and Jason-2 are retained, with normalization
    statistics estimated from 2012 onward rather than 2016 to use their longer records.
\end{itemize}

\begin{figure*}[t]
\centering
\includegraphics[width=0.49\linewidth]{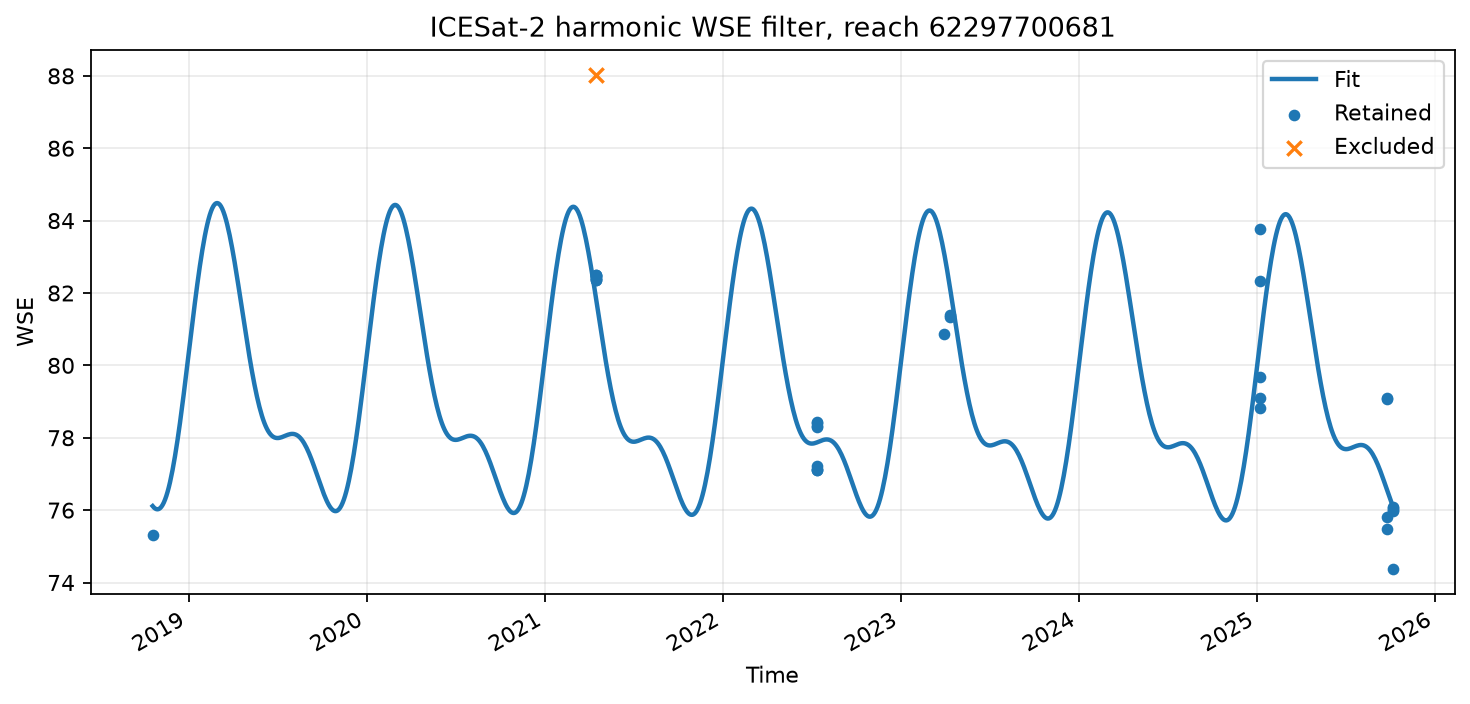}\hfill
\includegraphics[width=0.49\linewidth]{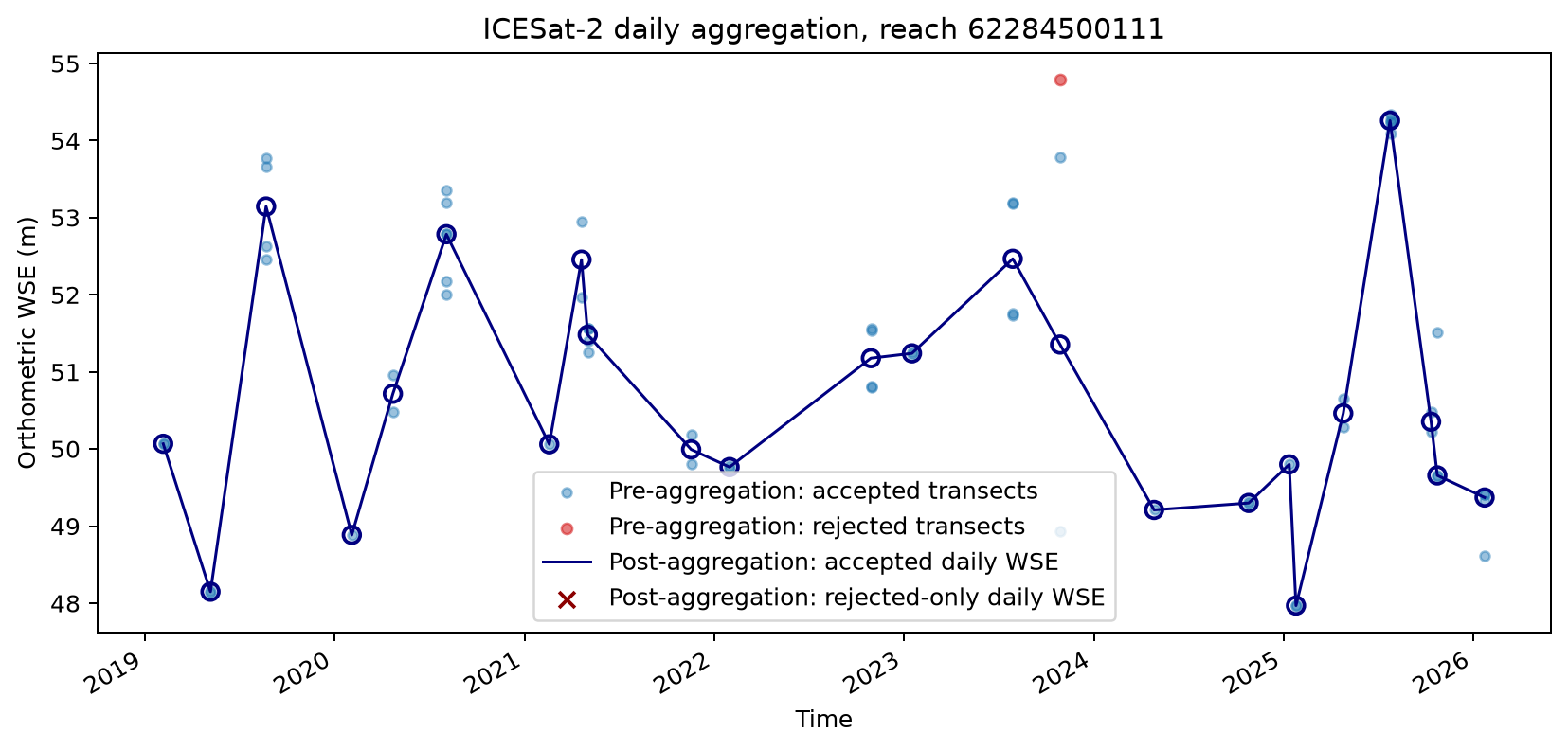}
\caption{
ICESat-2 harmonics filter (left) and aggregation of transect measurements per reach per day with accepted and rejected measurements (right).
}
\label{fig:icesat2-filters}
\end{figure*}

\subsubsection{ICESat-2.}

The filters remove about 4\% of measurements (of the subset of measurements that has been matched to a SWORD reach).

\begin{itemize}
    \item \textbf{Transect Matching.} The ATL22 contains measurements per transect, which is a contiguous
    portion of one ICESat-2 beam crossing a water body; land interruptions such
    as islands can split one beam pass into several transects. 
    Transects are joined to their nearest SWORD reach. Reservoir observations and
    transects shorter than 50~m are excluded.
    \item \textbf{Deviation from Harmonics.} For reaches with at least 10 transects, a two-harmonic model is fitted
    to transect orthometric heights and observations with residual modified
    Z-scores above $4.5$ are rejected.
    \item \textbf{Transect Aggregation.} Remaining transects are first aggregated within each reach-day and
    beam pass (source file and beam ID) using their median orthometric height.
    Reach-day WSE is the median of these beam-pass values, so a beam split into
    several transects does not receive extra weight.
    \item \textbf{Uncertainties.}
    We compute \texttt{wse\_u} as the leave-one-beam-pass-out
    jackknife standard uncertainty of the reach-day median.
    The ATL22 \texttt{ht\_stdev} field is provided as the median
    within-transect standard deviation of filtered short-segment heights, but 
    it should not be interpreted as a standard error of the transect mean. The output
    also retains the slope-fit RMSE, transect counts, beam-pass counts, and
    distinct-beam counts. 
\end{itemize}

\subsubsection{ANA in situ gauges}

\begin{figure*}[t]
\centering
\includegraphics[width=0.499\linewidth]{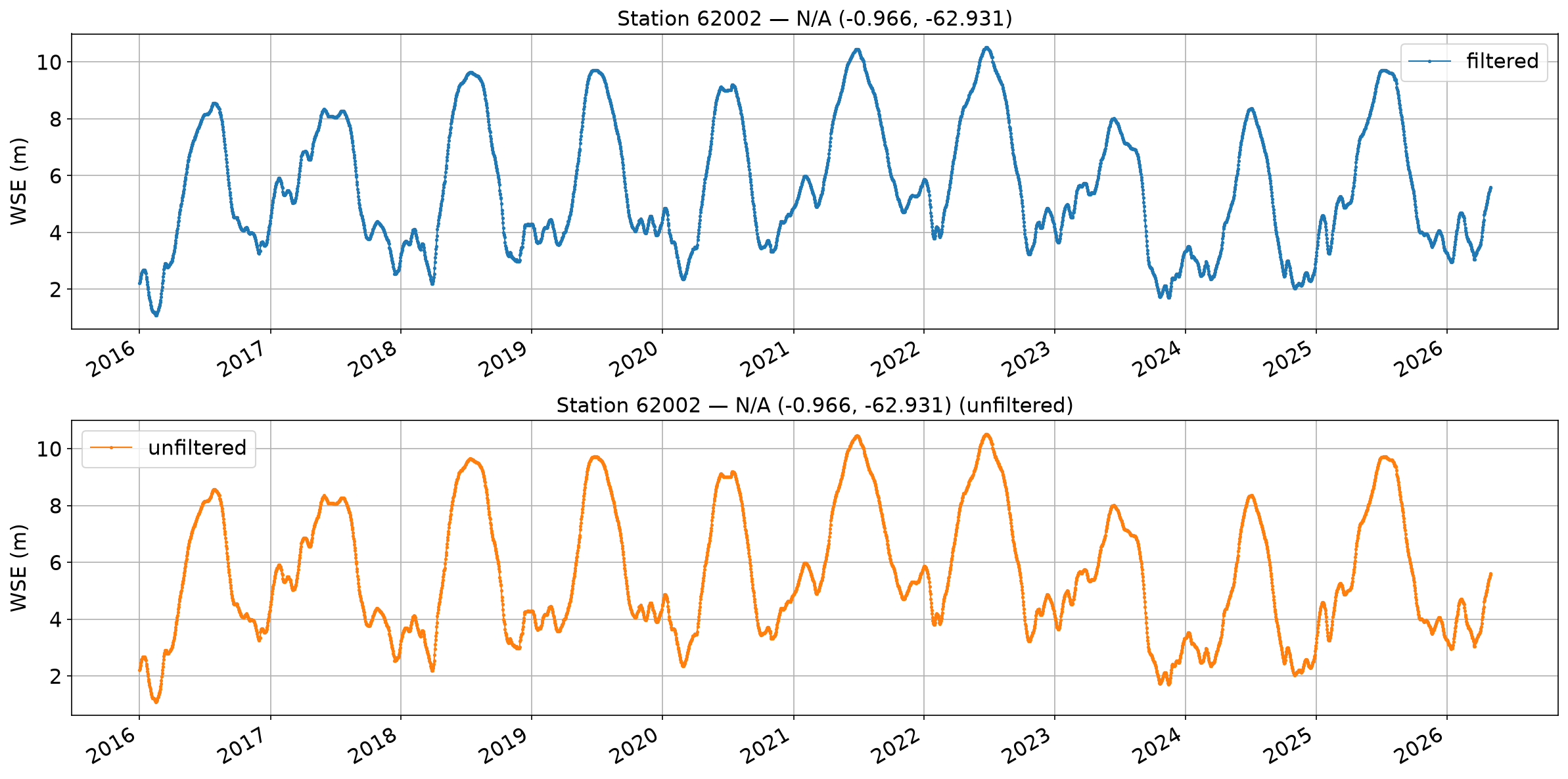}\hfill
\includegraphics[width=0.499\linewidth]{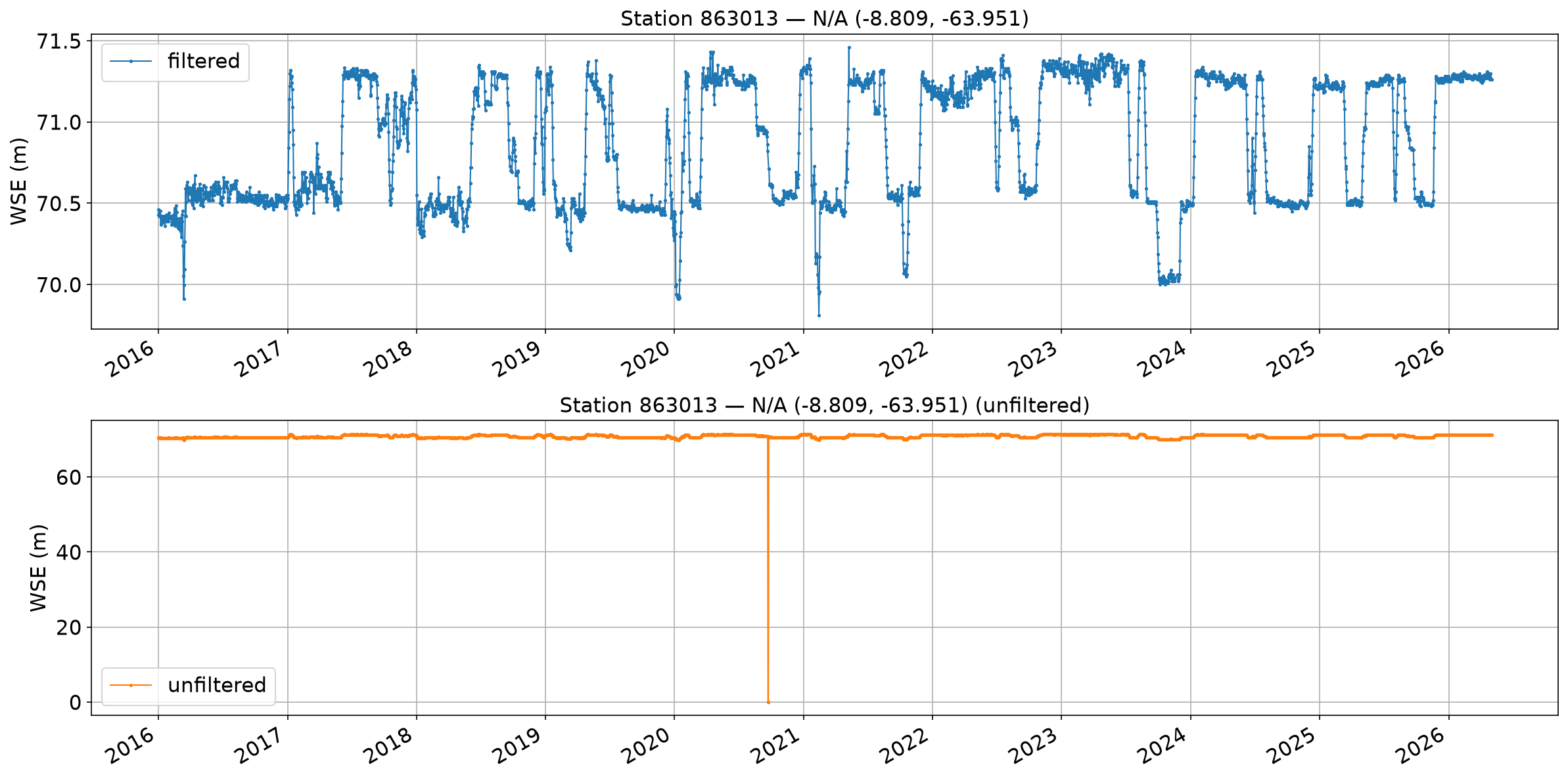}
\includegraphics[width=0.499\linewidth]{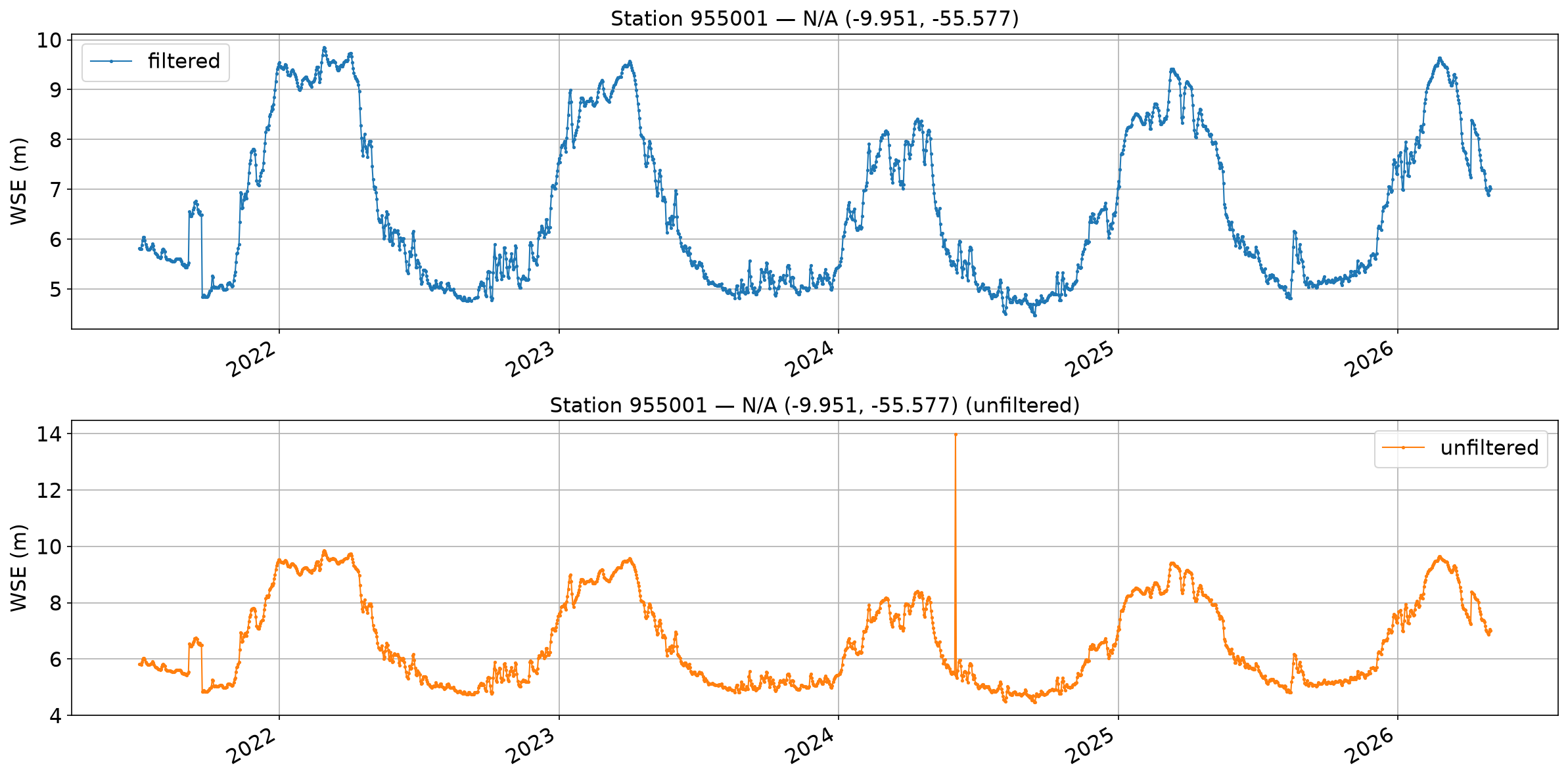}\hfill
\includegraphics[width=0.499\linewidth]{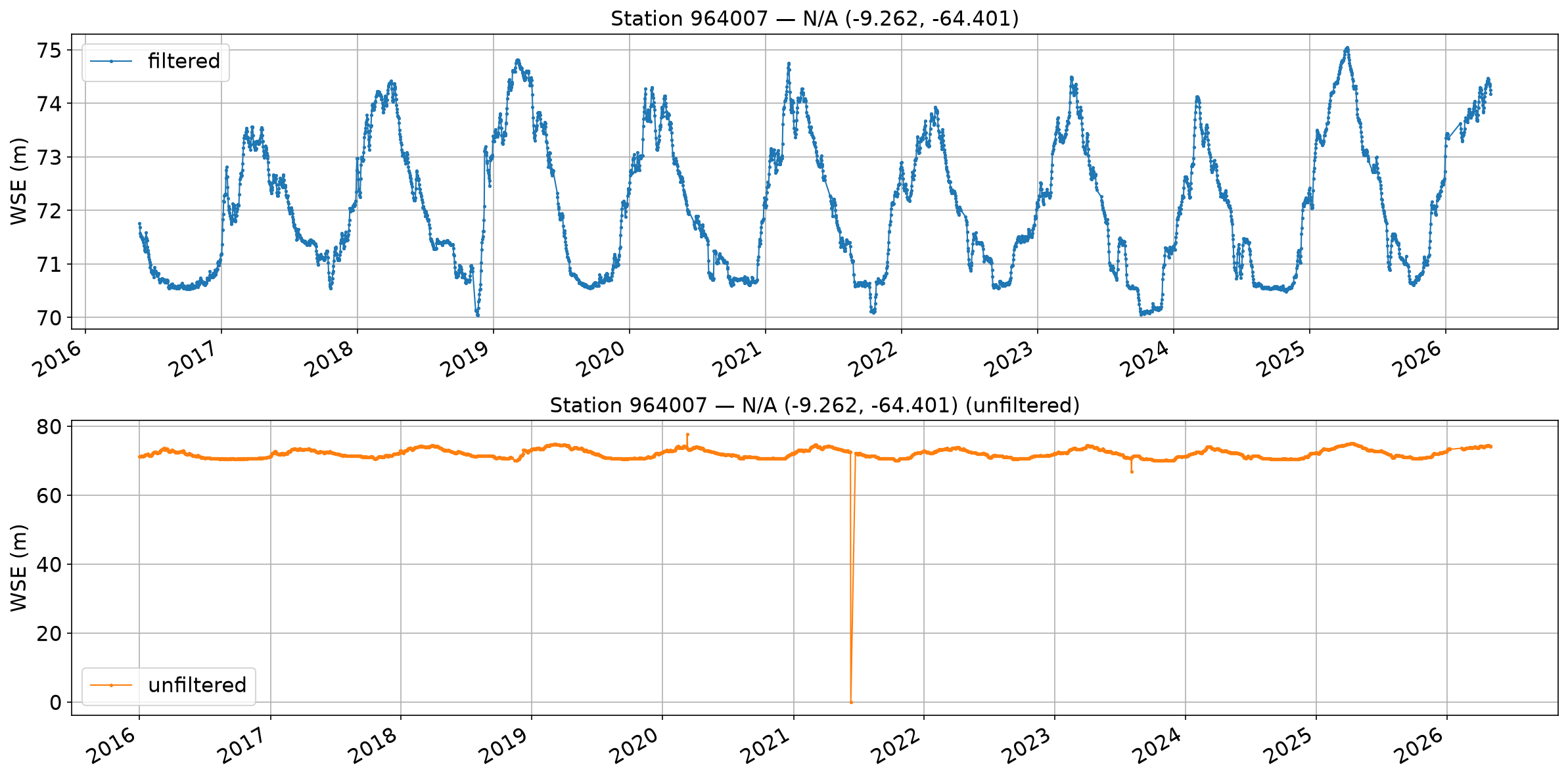}
\caption{
Filtered (blue) and unfiltered (orange) in situ gauge time series: a clean time series (top, left), a faulty time series that shows sudden jumps that clearly do not represent natural variations in river WSE (top, right), and two time series where automatic filters removed outliers (bottom, left and right).
}
\label{fig:insitu-filters}
\end{figure*}

Figure~\ref{fig:insitu-filters} show examples or in situ gauges before and after quality filters.
21\% of the 1000 timeseries are rejected automatically (10\% of all measurements), of the remaining 810, 589 have a SWORD reach match within 10~km, and of those, we retain 375 series.

\begin{itemize}
\item \textbf{ANA Timestamps.} are converted from Brazilian standard time (UTC-3),
to UTC and binned to calendar days. Multiple observations
within a day are represented by their median and converted from centimetres to
meters. 
\item \textbf{Uncertainty.} As ANA provides no per-observation measurement
uncertainty, we use the within-day population standard deviation of the
contributing readings as a data-derived uncertainty
measure.
    \item \textbf{Minimum Amount and Quality.} Stations must span at least two annual cycles (730 days) and contain at
    least 100 finite daily observations before filtering. 
    Repeated-value runs longer than three observations are removed.
    Negative observations are retained because gauge stage is referenced to
    a local datum.
    \item \textbf{Seasonality.} Daily series are
    linearly interpolated only to fit an STL seasonal--trend decomposition; the
    original observations are retained for filtering. Outliers are identified
    iteratively, for up to five STL fits, using a modified $z$-score of the
    residuals with a threshold of 4.5.
    \item \textbf{Manual Inspection.} An expert hydrology scientist manually inspects the remaining 581 time series and discards those that contain clear faulty measurements: 375 time series remain.
    \item \textbf{Datum Alignment.} Satellite sources report WSE w.r.t. a reference geoid~\cite{pavlis2012egm2008}, i.e., the elevation w.r.t. an approximate sea level. On the other hand, the in situ time series are measured in reference to a local gauge datum, for instance w.r.t. the local river bed bottom, which does not include the terrain elevation. Hence, before comparison, in all evaluations, we follow common practice in hydrology~\cite{halicki2026reachreg} to project the in situ measurements with a per-location fitted linear regression to the reference scale of the satellite data and/or predictions. 
    This operation is considered part of our evaluation benchmark, will be part of published evaluation code, and should be replicated by future authors that use this benchmark dataset. 
\end{itemize}

\begin{figure*}[t]
\centering
\includegraphics[width=0.332\linewidth]{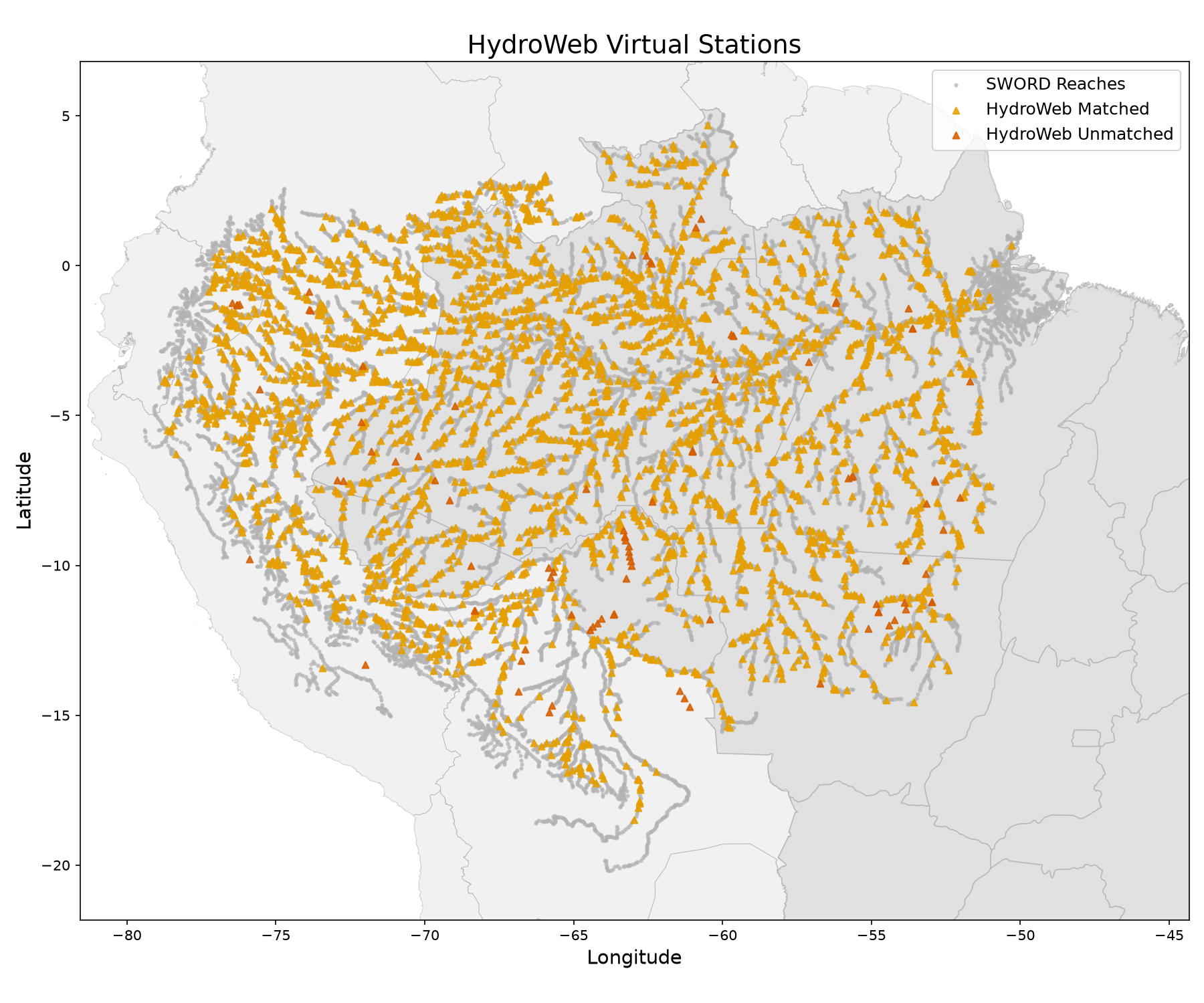}\hfill
\includegraphics[width=0.332\linewidth]{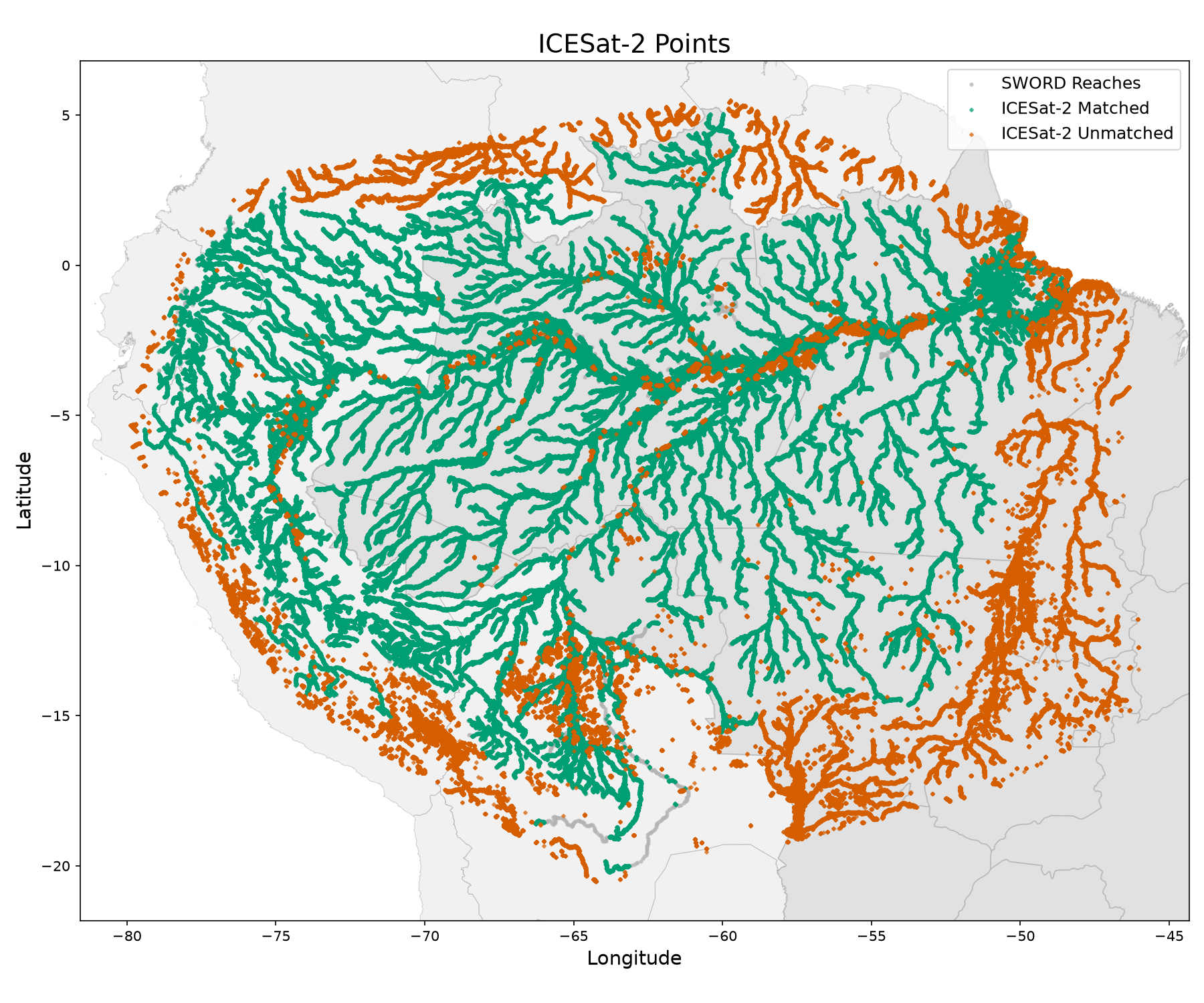}\hfill
\includegraphics[width=0.332\linewidth]{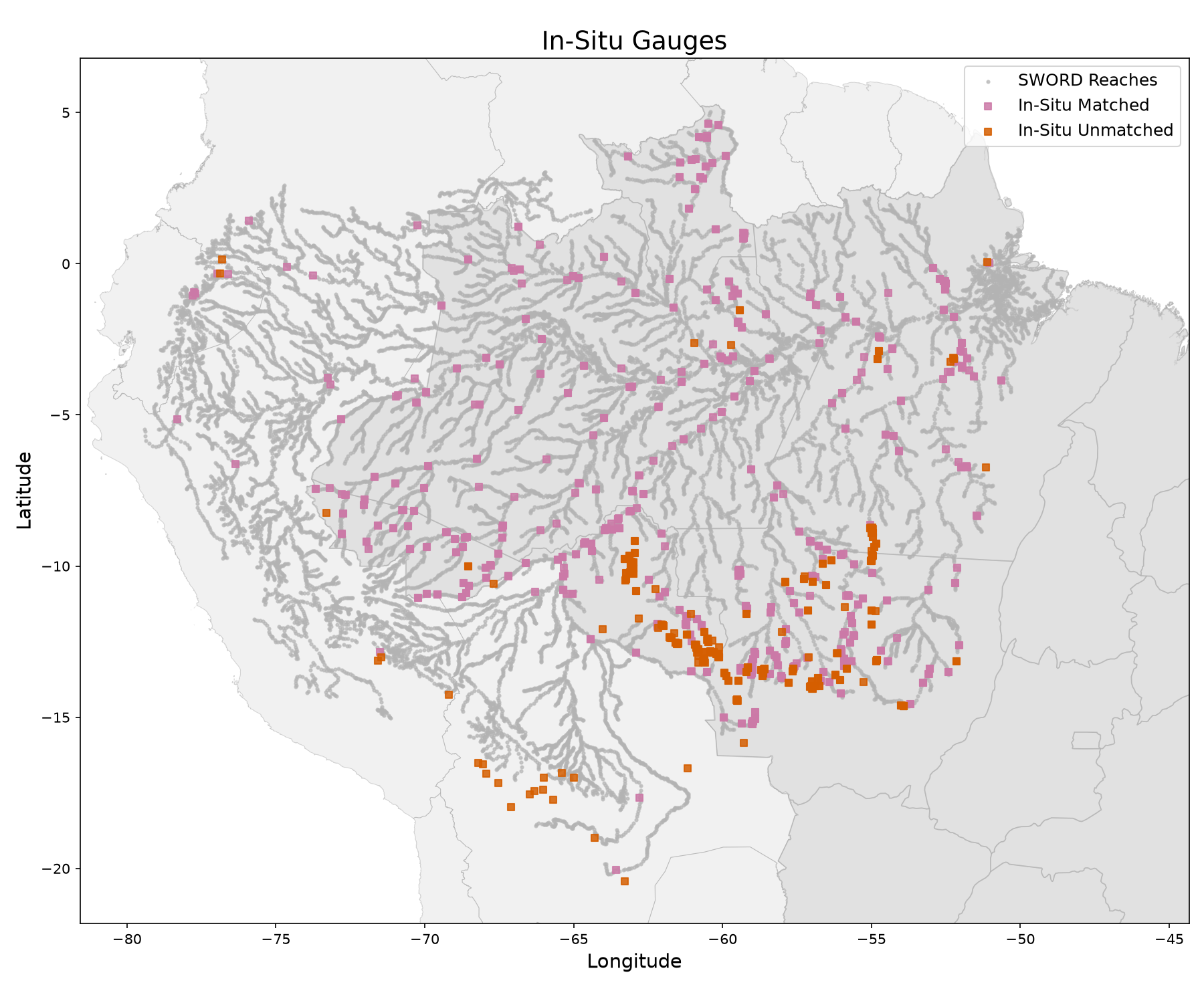}
\caption{Spatial distribution of data sources in \texttt{AmazonWSE}, matched (kept) and unmatched (discarded) to SWORD reaches within 10~km (\textit{before} quality filtering). From left to right: HydroWeb virtual stations, ICESat-2 transects, and ANA in situ gauges.}
\label{fig:matching-to-sword}
\end{figure*}

\begin{table*}[t]
\centering
\small
\begin{tabular}{@{}lll@{}}
\toprule
\textbf{Variable} & \textbf{Type} & \textbf{Description} \\
\midrule
\multicolumn{3}{@{}l}{\textit{Location variables (\texttt{location})}} \\
\addlinespace[2pt]
\texttt{location\_id}             & int64   & Unique site identifier
    (\texttt{cf\_role=timeseries\_id}) \\
\texttt{location\_quality\_flag}  & int8    & Site validity: 0 rejected, 1 accepted \\
\texttt{latitude}                 & float32 & Latitude (\texttt{degrees\_north}) \\
\texttt{longitude}                & float32 & Longitude (\texttt{degrees\_east}) \\
\texttt{sword\_reach\_id}         & int64   & Associated SWORD reach identifier \\
\texttt{reach\_distance\_m}       & float32 & Distance to the associated reach (m) \\
\addlinespace
\multicolumn{3}{@{}l}{\textit{Observation variables (\texttt{observation})}} \\
\addlinespace[2pt]
\texttt{observation\_location\_index}
                                  & int32   & Index into the \texttt{location} dimension \\
\texttt{time}                     & datetime64[ns] & Observation time (CF convention) \\
\texttt{wse}                      & float32 & Water-surface elevation (m) \\
\texttt{wse\_u}                   & float32 & Water-surface elevation uncertainty (m) \\
\texttt{quality\_flag}            & int8    & Observation validity: 0 rejected, 1 accepted \\
\bottomrule
\end{tabular}
\caption{NetCDF schema shared by all data sources. Variables are
grouped by their associated dimension.}
\label{tab:source-schema}
\end{table*}

\begin{table*}[t]
\centering
\small
\begin{tabular}{@{}lll@{}}
\toprule
\textbf{Group} & \textbf{Hyperparameter} & \textbf{Value} \\
\midrule
Architecture
    & Temporal model & Mamba-1 \\
    & Bidirectional; decoding order & true; time then flow \\
    & Embedding / SSM state dimension & 192 / 16 \\
    & Bidirectional SSM layers & 3  \\
    & Step rank; convolution width & 12; 4 \\
    & Dropout rate & 0.5 \\
    & Metadata combination; reinjection & elementwise addition; true \\
    & Output heads & satellite-specific \\
    & Tree embedding dimension $F$~\cite{shiv2019tree} & 4 \\
\midrule
Optimization
    & Loss & MSE \\
    & Optimizer & AdamW~\citeapp{loshchilov2018decoupled} \\
    & Learning rate; weight decay & $10^{-4}$; 0.05 \\
    & Batch size; maximum steps & 16; 100,000 \\
    & Warmup steps; gradient clipping & 1,000; 1.0 \\
    & Learning-rate schedule & reduce on plateau (factor 0.2, patience 5) \\
    & Validation interval; early-stopping patience & 500 steps; 20 checks \\
    & Early-stopping metric & RMSE on location masked HydroWeb \\
    & Seed & 43 \\
\midrule
Input subgraphs
    & Bin size; temporal bins & 24 hours; 91 (days) \\
    & Minimum / maximum measurements & 15 (training), 0 (eval) / 500 \\
    & Maximum spatial distance & 300 km \\
    & Upstream / downstream sampling & 0.75 / 0.25 \\
    & Main-upstream trunk proportion & 0.33 \\
    & Maximum upstream / downstream hops & 30 / 30 \\
    & WSE normalization & per-location mean and standard deviation \\
\midrule
Masking
    & Training strategies (probabilities) & location (0.9), random (0.1) \\
    & Mask ratio & 0.66 \\
\bottomrule
\end{tabular}
\caption{Hyperparameters used for the SSM model.
}
\label{tab:hyperparameters}
\end{table*}

\begin{figure*}[t]
\centering
\includegraphics[width=0.99\linewidth]{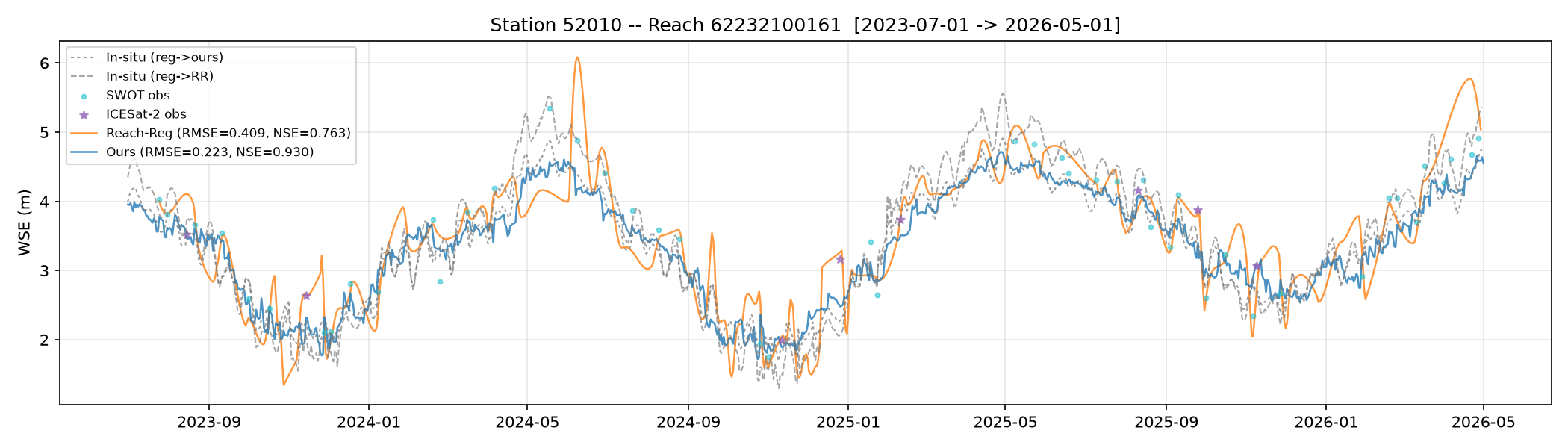}
\includegraphics[width=0.99\linewidth]{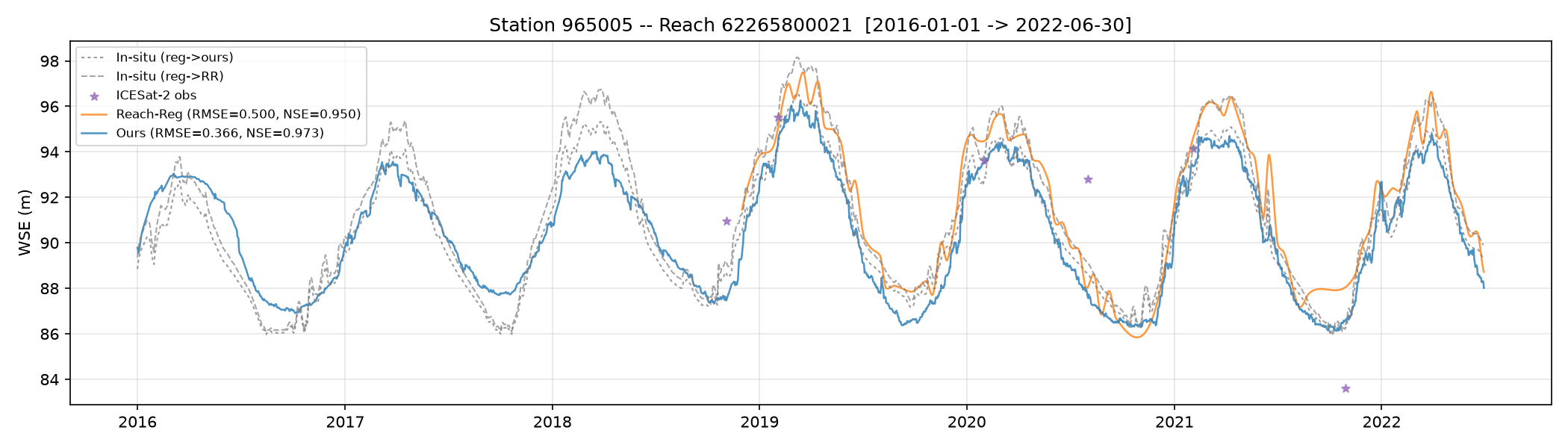}
\caption{
Example predictions of our method \includegraphics[height=10pt]{imgs/model/snake-emoji.pdf} compared to Reach-Reg on SWOT-era (top) and on hindcast (bottom, where Reach-Reg does not have enough input observations to predict the whole series).
The dotted lines represent the in situ gauge series linearly projected to the datum of our predictions and of Reach-Reg.
}
\label{fig:predictions}
\end{figure*}

\paragraph{Matching to SWORD Reaches.}

Figure~\ref{fig:matching-to-sword} shows matched and unmatched locations per source.
\begin{itemize}
    \item HydroWeb virtual stations, ICESat-2 transects, and ANA gauges are matched to
    their nearest SWORD reach using a 10\,km distance threshold (adapted from the 20~km threshold~\citet{halicki2026reachreg} uses for the Solimoes, which is part of the Amazon basin). Time series that are further than 10~km from the nearest SWORD reach are not used (though they are still included in the data files).
    \item SWOT is linked directly by the RiverSP reach identifier. 
\end{itemize}

\subsection{Storage Format}
\label{app:storage-format}

Each source is stored in a netcdf file with \texttt{h5netcdf} as a CF~1.13 indexed-ragged time series
\citepapp{eaton2025cf}, which means they are stored along a \texttt{location} dimension and an \texttt{observation} dimension (rather than a \texttt{time} dimension), where time is an additional variable.
This means only measured location--time pairs are stored, and missing (unobserved) location--time pairs do not take up storage space.
Table~\ref{tab:source-schema} summarizes the shared schema.
An example netcdf file per source is included in the \textit{Code and Data Supplement} along with the submission.

\paragraph{Indexing and Validity.}
The index variable declares \texttt{instance\_dimension=location}. Locations are
unique and sorted by \texttt{location\_id}; observations are sorted first by
location and then by time. Duplicate \texttt{(location\_id, time)} pairs are
forbidden. A quality flag of 0 marks a recorded but rejected value, a flag of 1
marks an accepted value, and a missing observation has no corresponding row.

\paragraph{Global Metadata.}
Each file records \texttt{schema\_version=1.0},
\texttt{featureType=timeSeries}, a non-empty \texttt{source}, and
\texttt{quality\_convention="quality\_flag: 0=rejected, 1=accepted"}.

\paragraph{Source-specific extensions.}
Sources may add variables without changing the two-dimensional organization:
\begin{itemize}
    \item \textbf{SWOT} adds observation-level \texttt{width},
    \texttt{width\_u}, \texttt{slope}, and \texttt{reach\_q\_b}, together with
    int32 \texttt{cycle\_id} and \texttt{pass\_id}.

    \item \textbf{ICESat-2} adds float32 \texttt{ht\_stdev} and
    \texttt{slope\_cross\_error} (m), as well as int32 counts of transects,
    beams, and their quality categories.

    \item \textbf{HydroWeb} adds observation-level \texttt{satellite},
    \texttt{orbit}, and \texttt{retracking\_algorithm} metadata.

    \item \textbf{ANA} adds the location-level string \texttt{river\_name}.
\end{itemize}

\subsection{Hyperparameters and Model Details}
\label{app:implementation}

Table~\ref{tab:hyperparameters} gives an overview of the used hyperparameters.
Importantly, since none of the models sees in situ data during training, and since our model makes source-aware predictions (i.e., with a separate linear decoding head per source), we make predictions as if for one of the training sources, also when we compare to in situ data. We de-normalize the prediction with the interpolated statistics from Figure~\ref{fig:mean-fields}: since the source that is being decoded might not have measurements hence also no statistics on the reach that is being predicted \textbf{and} statistics of the in situ gauge series should \textbf{not} be used. 

We find that decoding as HydroWeb works best when comparing to in situ gauge data, presumably since this source has long records of clean data.
This creates a discrepancy: we decode normalized HydroWeb predictions, and de-normalize them with interpolated SWOT statistics, but of all the tested combinations, this gave best results.

\subsection{Baselines}
\label{app:baselines}

All baselines use the splits, masking conventions, and sampled-subgraph settings, and hyperparameters from Table~\ref{tab:hyperparameters} unless stated otherwise.

\paragraph{Subgraphs vs full graph.}
We distinguish fixed-graph and sampled-subgraph settings. Fixed graphs use a
basin-wide time--node grid with stable reach identities, following the
transductive protocols of GRIN, SPIN-H, and ImputeFormer
\citep{cini2022grin,marisca2022spin,nie2024imputeformer}. Sampled sub-graph windows instead contain the same local river neighborhoods as our method (cfr. Table~\ref{tab:hyperparameters} and Figure~\ref{fig:subgraph}).
SPIN-H and IGNNK explicitly support subgraph sampling \citep{marisca2022spin,wu2021ignnk}, while for GRIN and ImputeFormer we run the subgraph version as adaptations rather than replications of their published protocols.
KITS is closest to its published increment-training setting when run on a fixed
graph with virtual nodes \citep{xu2025kits}.

\paragraph{Source representation.}
We consider two representations of co-located sources (SWOT, HydroWeb, ICESat-2): the \emph{channels} mode assigns each source a separate value (in a separate input channel), input
mask, and target mask, preserving source identity and allowing one source to
remain visible while another is held out. 
The \emph{merged} mode first normalizes sources with common SWOT-derived reach
statistics; after masking, visible source values are averaged into one scalar.
Thus, held-out observations never enter the aggregate, but source identity and
source-specific biases are removed.

Models are never trained on in situ data. At in situ gauges, we use the training source predictions and dense SWOT-derived reach statistics. IGNNK and
KITS use \textit{merged} inputs because they are scalar-WSE models~\citep{wu2021ignnk,xu2025kits}. For ImputeFormer, GRIN, and
SPIN-H we implement \textit{channel} variants, even though these methods were primarily evaluated with scalar targets~\citep{nie2024imputeformer,cini2022grin,marisca2022spin}.
We also tested representing each source--location pair as a separate node (allowing thus several nodes for the same reach), but it gave worse performance.

\paragraph{Non-graph controls.}
The non-parametric kNN baseline uses 91-day windows, ten spatial neighbors, and
equal spatial and temporal weights. It fits a dense field separately per input source and outputs the mean of the per-source predictions.
A temporal-only, one-layer bidirectional LSTM inspired by BRITS-I~\citep{cao2018brits} 
uses the \textit{channels} mode and 365-day per-reach windows, without graph information.

\paragraph{Graph-imputation models.}
For fixed-graph runs we compared 7-, 14-, and 28-day fixed windows (smaller than subgraph runs to fit in memory), as well as \textit{merged} vs. \textit{channels} mode, and chose the setting that gave best RMSE on validation samples.
We had to reduce the hidden dimension size of some baselines to fit their training in our GPU memory.
We reuse settings from the original publications and/or default settings in the published repositories where applicable.

GRIN \citep{cini2022grin} uses one recurrent graph-imputation layer, hidden size
64, kernel size 2, and decoder order 1. Its fixed and sampled variants use
\textit{channels} and windows of 28 and 91 days and feed-forward widths of 128 and 64,
respectively. SPIN-H \citep{marisca2022spin} uses five layers, hidden size 32,
$\eta=3$, and two message-passing layers. Its fixed variant uses 28-day \textit{merged}
inputs, latent size 128, and four heads; its sampled variant uses 91-day \textit{channels},
latent size 64, and two heads.
IGNNK~\citep{wu2021ignnk} uses hidden size 192, first-order diffusion, and \textit{merged} windows of 28 days (fixed graph) or 91 days (sampled subgraph). Its
graph convolutions reconstruct each time step from visible neighbors without
explicit temporal dynamics. KITS~\citep{xu2025kits} uses hidden size 64, a 0.3
virtual-node ratio, unit cycle-consistency weight, and \textit{merged} windows of 7 days
(fixed) or 91 days (sampled). Training-time virtual nodes are connected around
random one-hop neighborhoods and removed before evaluation.
ImputeFormer \citep{nie2024imputeformer} uses input, node-embedding, and
feed-forward dimensions of 64, 96, and 256; three layers; four temporal heads;
rank 8; dropout 0.1; and Fourier-loss weight 0.01. Its fixed variant uses a
14-day window and its sampled subgraph variant uses a 91-day window, both with source \textit{channels}.

\paragraph{Repositories.}
For GRIN, SPIN-H and the BRITS-I inspired bidirectional LSTM we use the implementations of \texttt{torch-spatiotemporal}.\footnote{\url{https://github.com/TorchSpatiotemporal/tsl}}
ImputeFormer\footnote{\url{https://github.com/tongnie/ImputeFormer}}, IGNNK\footnote{\url{https://github.com/Kaimaoge/IGNNK}} and KITS\footnote{\url{https://github.com/Sam1224/KITS}} are adapted from their published repositories.

\paragraph{Baselines that were not included.}
GgNet~\citep{de2024ggnet} is excluded because it assumes dynamic covariates available
at target reaches, unlike our sparse observations of the WSE target itself.
Diffusion imputers such as PriSTI and FastSTI
\citep{liu2023pristi,cheng2024faststi} are deferred pending a matched,
deterministic extreme-sparsity protocol. Forecasting-only methods are excluded
because they provide neither an imputation objective nor a corresponding masking
protocol.

\paragraph{Reach-Reg.}
Reach-Reg is a recent method that leverages the SWOT measurement geometry to fit a chain of linear regressions (orthogonal distance regressions) to propagate measurements between consecutive reaches. 
Subsequently, a time lag is estimated with the Manning formula for river velocity, and aggregation and interpolation then yield a daily WSE.
The approach is applied either to combined SWOT, S3 and S6 measurements from the DAHITI dataset~\cite{schwatke2015dahiti} or to SWOT RiverSP measurements only.
We use the official Reach-Reg implementation.\footnote{\url{https://github.com/MichalHalicki4/Reach-Reg/}, fetched at commit \texttt{59aa780}.}

We group the 375 gauges into 107 unique river segments (sequences of river reaches formed by traversing downstream and upstream from the target gauge, choosing the upstream branch with highest accumulated flow).
We run Reach-Reg with the settings of \citet{halicki2026reachreg} on SWOT RiverSP data (version D, as our model and baselines) retrieved and filtered from the HydroCron API with the settings and filters proposed by Reach-Reg.
We compare linear with Akima interpolation and find that it does not make a significant difference.
A prediction is not made for all gauges, since Reach-Reg does not make a prediction when quality filters leave insufficient neighboring stations for propagation or when propagation and path-error filtering leave no valid predictions.
The authors achieve a better accuracy when running Reach-Reg on timeseries from the DAHITI dataset, combining S3, S6 and SWOT~\cite{schwatke2015dahiti}, rather than on SWOT only, but since SWOT timeseries for only a small part of reaches in the Amazon have been included in DAHITI as of yet, this further reduces coverage by over \%50.

We next adapt Reach-Reg to make hindcast predictions for the available gauges (i.e., for the period from 2016-01 to 2022-06-30 for which no SWOT measurements are available).
We reuse the fitted orthogonal distance regression weights and Manning parameters that resulted from the prediction that was just described for SWOT RiverSP data.
We then inject historical DAHITI observations into the matching (and fitted) SWOT-era station slots and propagate them according the fitted and chosen parameters.
We increase the maximum cumulative ODR path-error, defined as the sum of the ODR RMSEs along a propagation path, from 10 to 50\,m. We also increase the per-link ODR-RMSE threshold used to choose a direct link versus an alternative path from 0.2 to 1.0\,m. We retain a propagated observation only when its cumulative path-error is below $\max(0.2 \cdot \text{WSE amplitude}, 1.5 \cdot \text{median}(\text{cumulative path-error}))$. These relaxations accommodate the longer regression chains needed in the much sparser hindcast setting.
It should be noted that this evaluation takes Reach-Reg out of the context it was developed for, and that results are hence only indicative.

\subsection{Predictions}
\label{app:predictions}

Figure~\ref{fig:predictions} shows examples of predicted time series.

\bibliographystyleapp{aaai2027}
\bibliographyapp{aaai2027}

\end{document}